\documentclass{article} 
\usepackage{iclr2026_conference,times}

\usepackage{amsmath,amsfonts,bm}

\def\eqref#1{equation~\ref{#1}}

\def\1{\bm{1}}

\DeclareMathAlphabet{\mathsfit}{\encodingdefault}{\sfdefault}{m}{sl}
\SetMathAlphabet{\mathsfit}{bold}{\encodingdefault}{\sfdefault}{bx}{n}

\newcommand{\softmax}{\mathrm{softmax}}

\usepackage{hyperref}
\usepackage{url}
\usepackage{multirow}

\usepackage{wrapfig}
\usepackage{multicol}
\usepackage{algorithm}
\usepackage{algpseudocode}

\usepackage{titletoc}
\usepackage[utf8]{inputenc} 
\usepackage[T1]{fontenc}    
\usepackage{amsmath}
\usepackage{graphicx} 
\usepackage{hyperref}       
\usepackage{cleveref}      
\usepackage{url}            
\usepackage{booktabs}       
\usepackage{tikz}           
\usepackage{amsfonts}       
\usepackage{nicefrac}       
\usepackage{microtype}      
\usepackage{xcolor}         
\usepackage{subcaption}

\usepackage{graphicx}     
\usepackage{subcaption}   
\usepackage{caption}
\usepackage{enumitem}

\title{On the Predictive Power of Representation Dispersion in Language Models}

\author{%
  Yanhong Li \\
  University of Chicago \\
  \texttt{yanhongli@uchicago.edu} \\
  \And
  Ming Li \\
  University of Maryland \\
  \texttt{minglii@umd.edu} \\
  \And
  Karen Livescu \\
  Toyota Technological Institute at Chicago \\
  \texttt{klivescu@ttic.edu} \\
  \And
  Jiawei Zhou \\
  Stony Brook University \\
  \texttt{jiawei.zhou.1@stonybrook.edu} \\
}

\author{
\hspace{-2mm} \textbf{Yanhong Li}$^{1}$ \quad
\textbf{Ming Li}$^{2}$ \quad
\textbf{Karen Livescu}$^{3}$ \quad
\textbf{Jiawei Zhou}$^{4}$ \\
$^{1}$Allen Institute for AI \quad
$^{2}$University of Maryland \quad
$^{3}$Toyota Technological Institute at Chicago \\
$^{4}$Stony Brook University \\
\texttt{yanhongl@allenai.org \,\, jiawei.zhou.1@stonybrook.edu}
}

\iclrfinalcopy 
\begin{document}

\maketitle

\begin{abstract}
We show that a language model’s ability to predict text is tightly linked to the \emph{breadth} of its embedding space: models that spread their contextual representations more widely tend to achieve lower perplexity. Concretely, we find that \emph{representation dispersion}—the average pairwise cosine distance among hidden vectors—strongly and negatively correlates with perplexity across diverse model families (LLaMA, Qwen, and others) and domains (Wikipedia, news, scientific abstracts). Beyond illustrating this link, we show how dispersion can be leveraged for a range of practical tasks—without requiring labeled data. 
First, measuring dispersion on unlabeled text allows us to rank examples by difficulty and identify hard slices in new domains, offering a data‐efficient tool for screening and prioritizing models before full evaluation.
Next, we find that identifying layers with higher dispersion pinpoints the best representations for retrieval‐based methods such as $k$NN‐LM, bypassing exhaustive layer‐by‐layer searches. Finally, we integrate a simple “push‐away” objective into training, which increases dispersion in both single‐domain and cross‐domain scenarios and directly improves perplexity in each.\footnote{Code is available at \url{https://github.com/yanhong-lbh/rep_dispersion}.}
\end{abstract}

\section{Introduction}
\label{sec:intro}

Large language models can perform remarkably well on tasks ranging from text
completion to code generation.  Yet their \emph{embedding geometry} often exhibits
signs of anisotropy or rank collapse, whereby hidden states lie in a narrow cone or occupy a low‐dimensional subspace \citep{DBLP:conf/emnlp/Ethayarajh19,DBLP:conf/iclr/GaoHTQWL19,DBLP:conf/emnlp/LiZHWYL20}.
Although this geometry has been posited to limit expressive power, \emph{how} precisely it connects to auto-regressive text generation remains less clear.

\begin{figure}[t]
    \centering
    \begin{minipage}{0.64\textwidth}
        \centering
        \includegraphics[width=\textwidth]{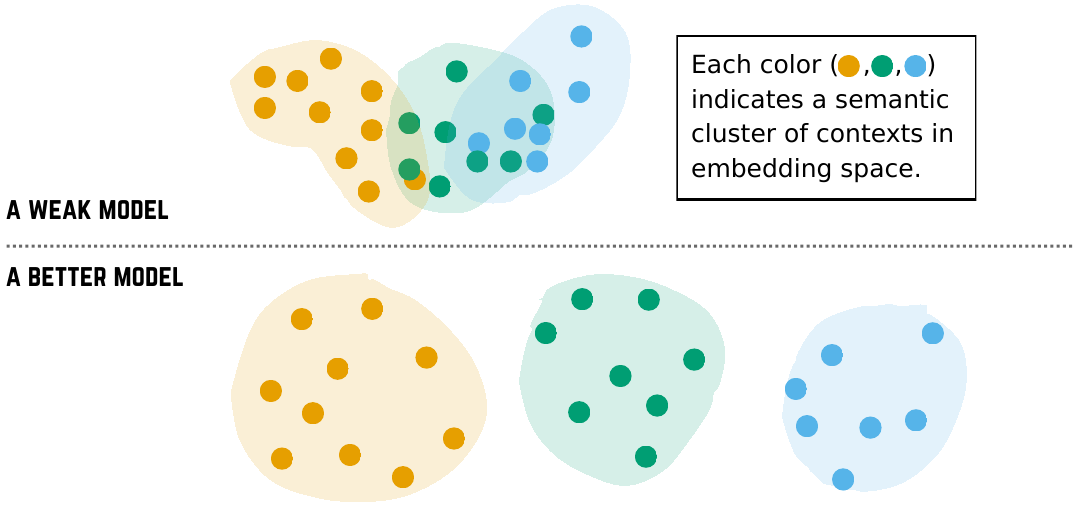}
        \caption{\textbf{Representation geometry in a weak model vs.\ a better model.} In a weak model (top), final‐layer embeddings for similar contexts are compressed into tight clusters, limiting discriminative power. In a better model (bottom), embeddings are widely dispersed—even within semantically related clusters—leading to more confident next‐token predictions.}
        \label{fig:teaser-a}
    \end{minipage}
    \hfill
    \begin{minipage}{0.33\textwidth}
        \centering
        \includegraphics[width=\textwidth]{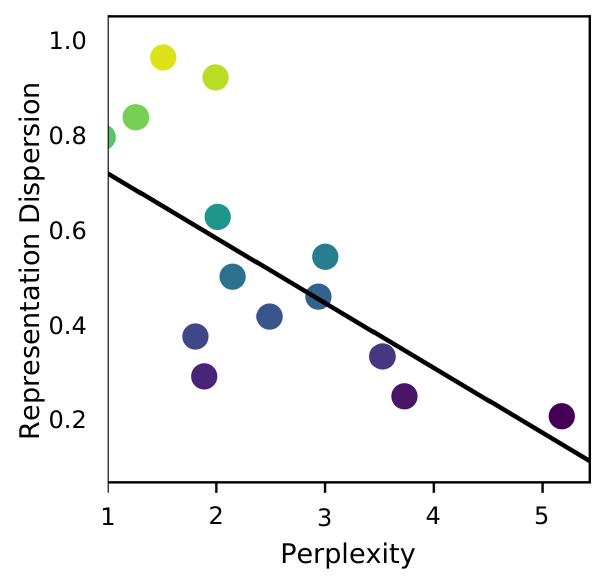}
        \caption{\textbf{An Illustrative Empirical Trend.} 
        Across models/datasets, higher 
        dispersion correlates with lower perplexity.}
        \label{fig:teaser-b}
    \end{minipage}
\end{figure}

In this paper, we present empirical evidence that a model’s ability to predict
text is tightly linked to the \emph{breadth} of its embedding space. Intuitively, as illustrated in
\Cref{fig:teaser-a}, weaker models compress contexts into tight clusters, whereas stronger models separate these contexts —\emph{even
semantically similar ones}—more broadly.  This broader geometry yields clearer distinctions
in the latent space, enabling sharper (lower‐entropy) next‐token predictions. 
Concretely, we quantify \textbf{representation dispersion} at \emph{any chosen layer} as the average pairwise cosine distance of its hidden vectors. Unless specified otherwise, our empirical sections use the final layer, and we show that higher dispersion consistently predicts \emph{lower perplexity} (\Cref{fig:teaser-b}).

Beyond revealing this fundamental link, representation dispersion offers
\emph{practical} benefits:
\begin{itemize}[itemsep=1pt,topsep=1pt,parsep=1pt,leftmargin=2em]
    \item \textbf{Label-free diagnostics.}
          Dispersion measured on \emph{unlabeled} text monotonically tracks correctness for a fixed model and dataset, enabling label-free \emph{ranking} of examples by difficulty and discovery of hard slices
          (\S\ref{sec:pred-downstream}).
    \item \textbf{Model selection.}
          Among multiple pretrained or fine-tuned variants within the same model family, larger dispersion provides a coarse, inexpensive signal that helps identify clearly under-adapted checkpoints and prioritize promising ones for full evaluation
          (\S\ref{sec:embedding-dispersion-modelsel}).
    \item \textbf{Layer selection for retrieval augmentation.} Although the first two applications exploit the final hidden state, dispersion is equally informative inside the network. In $k$NN-LM, choosing the hidden layer with the highest dispersion
          yields the best perplexity, providing an unsupervised shortcut to
          sub-layer selection (\S\ref{sec:layer-selection}).
    \item \textbf{A training signal.}
          Encouraging higher dispersion through an auxiliary
          “push-away” loss directly improves perplexity for both single‐domain and cross‐domain scenarios  (\S\ref{sec:rd-regularization}). 
\end{itemize}


Overall, we show that a model’s embedding geometry—captured by the simple statistic of \emph{average pairwise distance}—serves as a robust indicator of predictive quality. 
By quantifying and encouraging this broader
geometry, we gain both conceptual insight and practical benefits, and we hope
this perspective fosters new avenues for improving model robustness and
interpretability.




\begin{figure}[t]
    \centering
    \includegraphics[width=\textwidth]{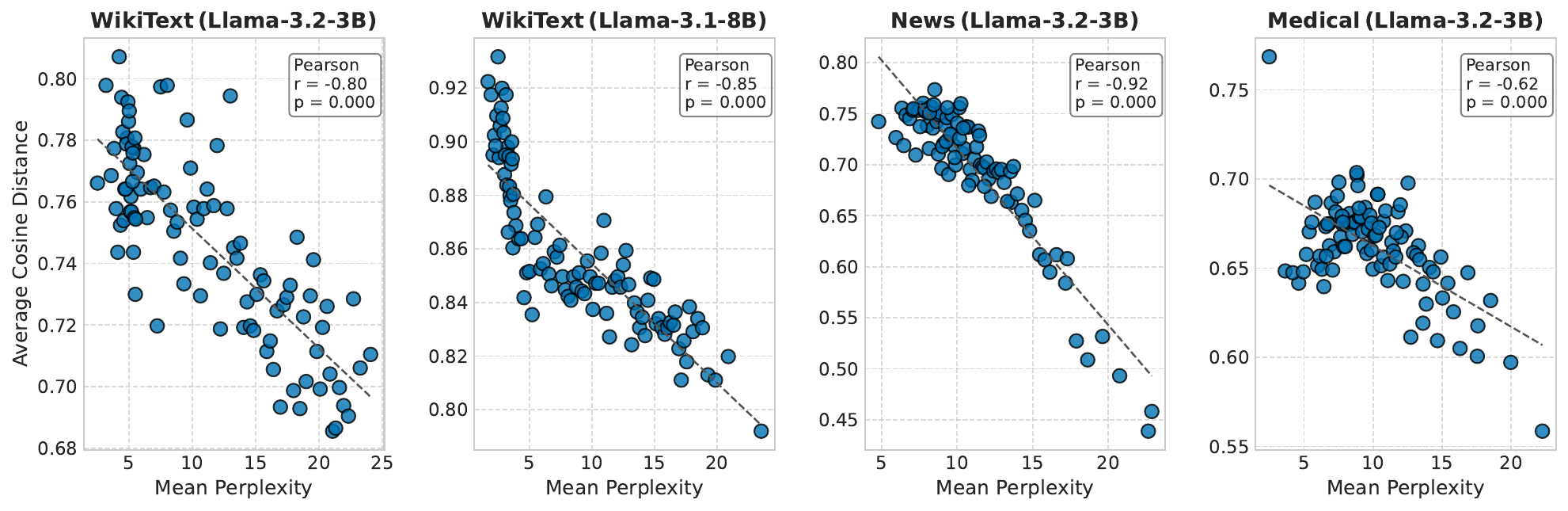}
    \caption{Sequence-level perplexity vs.\ embedding dispersion across different domains and model sizes.
Each point represents a bin of text segments with the x-axis showing mean sequence-level perplexity and the y-axis showing average pairwise cosine distance of final-layer embeddings.}
    \label{fig:seq_len_main}

    \vspace{1em}

    \includegraphics[width=\textwidth]{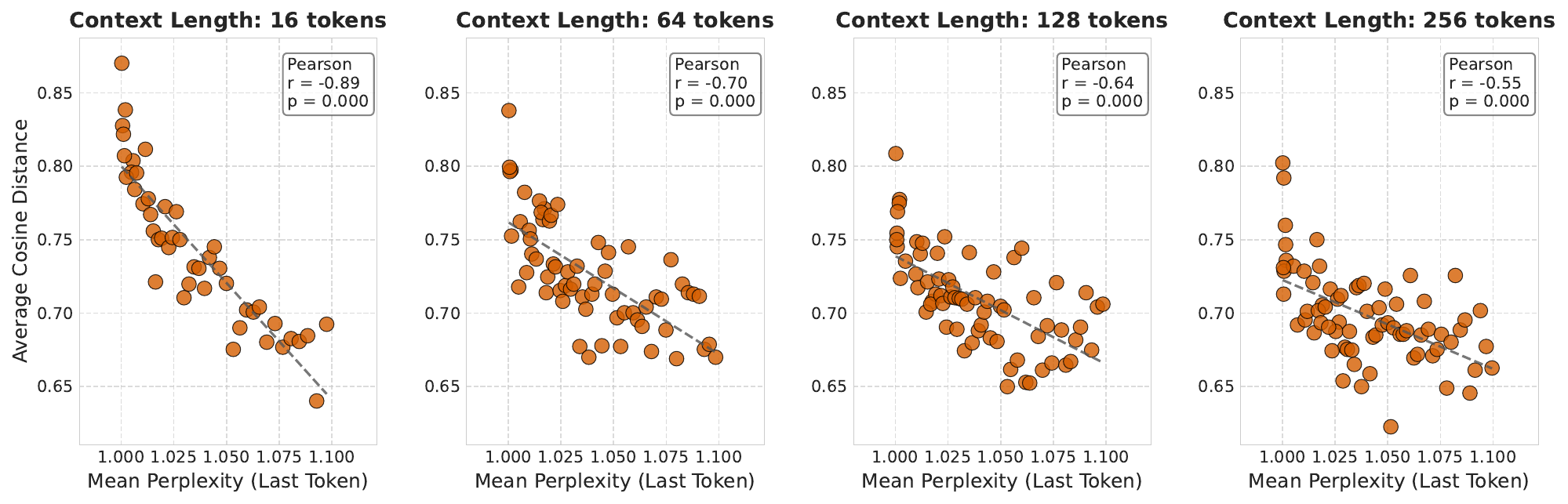}
    \caption{Last-token perplexity vs.\ embedding dispersion at varying context lengths using LLaMA-3.2-1B.
Each point represents a bin of text segments with the x-axis showing mean last-token perplexity and the y-axis showing average pairwise distance.}
    \label{fig:last_token_main}
\end{figure}

\section{Empirical Analysis of Representation Geometry}
\label{sec:analysis}

We begin by describing how we measure representation geometry and perplexity (\S\ref{subsec:measurement-setup}). We then present three key global observations that characterize how embedding dispersion evolves with perplexity, across layers, and under fine-tuning (\S\ref{subsec:global-observations}). Finally, we zoom in on semantic clusters (\S\ref{subsec:semantic-dispersion}) to examine how dispersion behaves among closely related contexts.

\subsection{Measurement Setup}
\label{subsec:measurement-setup}
\paragraph{Contextual Representations.}
Following standard autoregressive conventions, let a language model with parameters \(\theta\) assign probability to a token sequence \((x_1, x_2, \ldots, x_N)\) via
$p_{\theta}(x_1, x_2, \ldots, x_N) = \prod_{n=1}^N p_{\theta}\bigl(x_n \mid x_{<n}\bigr)$,
where \(x_{<n} = (x_1,\ldots,x_{n-1})\). 
In contemporary Transformer-based models, each partial sequence \(x_{<n}\) is mapped to an internal \emph{context vector} \(\mathbf{h}_n\in\mathbb{R}^d\) by a function \(f_\theta(\cdot)\). The next-token distribution is then given by 
$p_{\theta}\bigl(x_n \mid x_{<n}\bigr) = \softmax\!\bigl(W_o\, \mathbf{h}_n\bigr)$,
where \(W_o\in\mathbb{R}^{|V|\times d}\) projects the representation \(\mathbf{h}_n\) into the vocabulary space \(V\). 
For measuring \emph{representation dispersion}, we focus on a chosen final-layer vector (e.g., \(\mathbf{h}_N\) for the full sequence), by default, as the representation of each text sample.

\paragraph{Measuring Representation Dispersion.}
To measure how well the model separates text samples in its embedding space, we first choose a particular layer or sub-layer whose representation we wish to examine (e.g., the final hidden state, the output after the final attention block, or the second-to-last block). Concretely, we sample \(N\) text segments from a dataset and pass each segment through the model to extract the corresponding representation \(\mathbf{E}_i \in \mathbb{R}^d\) from the chosen sublayer, for \(i=1,\ldots,N\). We then compute the \textbf{average pairwise cosine distance} of these representations:
\begin{equation}\label{eq:avg-dist}
    \overline{D} \;=\; \frac{1}{\binom{N}{2}}
    \sum_{1 \le i < j \le N}
    \Bigl[1 - \frac{\mathbf{E}_i \cdot \mathbf{E}_j}{\|\mathbf{E}_i\|\;\|\mathbf{E}_j\|}\Bigr].
\end{equation}
This quantity reflects how ``spread out'' the embeddings are, with higher values indicating greater separation among representations.

\subsection{Global Observations on Representation Dispersion}
\label{subsec:global-observations}

In this section, we examine how the model’s representation space behaves across a broad sample of text segments, highlighting how perplexity, layer depth, and fine-tuning each affect embedding dispersion.

\paragraph{(1) Perplexity vs.\ Representation Dispersion.}
Our first finding connects a model's \emph{sequence-level perplexity} to how spread out its contextual embeddings are.\footnote{%
  We define the sequence-level perplexity of a text segment \(x_{1:L}\) of length \(L\) as:
  \[
    \mathrm{ppl}(x_{1:L}) \;=\; \exp\!\biggl(
      -\frac{1}{L} \sum_{t=1}^{L} \log p_{\theta}(x_t \,\mid\, x_{<t})
    \biggr),
  \]} We randomly select 100,000 text segments of 512 tokens, compute their perplexities, and also measure the \textbf{average pairwise distance} of their final-layer embeddings. We sort by perplexity and group segments into bins, and for each bin recording its mean perplexity and mean pairwise distance.

\Cref{fig:seq_len_main} reveals a strong \textbf{negative correlation}: Segments with \emph{lower} perplexity have \emph{more dispersed} embeddings, whereas those with \emph{higher} perplexity show \emph{more compressed} embeddings. This trend appears across multiple model families (Llama, Phi, Mistral, Qwen) and diverse text domains (e.g.\ Wikitext-103, CNN Daily News, PubMed summarization). Full visualizations are provided in \Cref{subsec:all_vis}.

We also verify that the relationship holds at a finer granularity by focusing on \emph{last-token perplexity}, shown in \Cref{fig:last_token_main}. Across context lengths of 16, 64, 128, and 256 tokens, we observe the same negative correlation trend.
Intuitively, this negative correlation appears because contexts with more confident next‐token predictions (low perplexity) end up pushed into more distinct regions of the embedding space, whereas harder‐to‐predict contexts remain more compressed.

\begin{figure}[t]
    \centering
    \includegraphics[width=\linewidth]{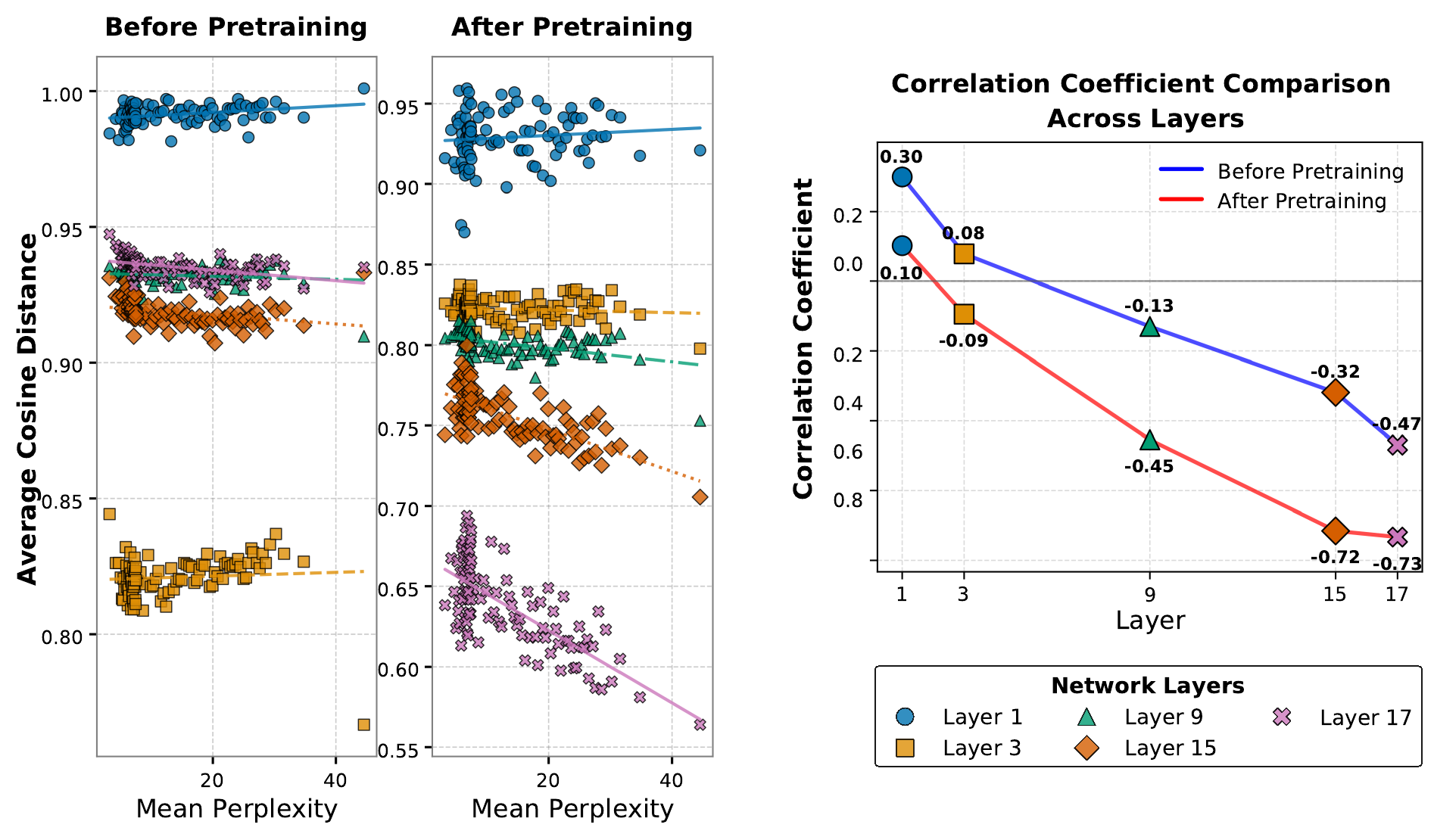}
    \caption{Layer-wise embedding separation in LLaMA-3.2-1B, shown across different perplexity bins (left panel) 
    and the corresponding correlation coefficients (right panel). 
    Note that the negative correlation between perplexity and embedding separation becomes more pronounced 
    as we move to deeper layers (see right panel).}
    \label{fig:layerwise-sep}
\end{figure}

\paragraph{(2) Layer-Wise Patterns.}
We next assess how this relationship unfolds across layers. Collecting embeddings from multiple intermediate layers (e.g.\ Layers 1, 3, 9, 13, 17) and replicating the above procedure, we find that \emph{the negative correlation strengthens in deeper layers} (as illustrated in the right panel of \Cref{fig:layerwise-sep}). Early layers do not show a clear correlation, likely because they capture lower-level lexical features rather than global predictive cues. Deeper layers exhibit \emph{more pronounced} embedding distance differences between easier-to-predict and harder-to-predict samples. Figure~\ref{fig:layerwise-sep} illustrates this layer-wise progression for a representative LLaMA-3.2-1B model. Additionally, comparing models before and after pretraining indicates that the negative correlation emerges primarily after the model has been trained to predict tokens, pointing to a learned representational structure.

\begin{figure}[h]
    \centering
    \begin{minipage}{0.31\textwidth}
        \centering
        \includegraphics[width=\linewidth]{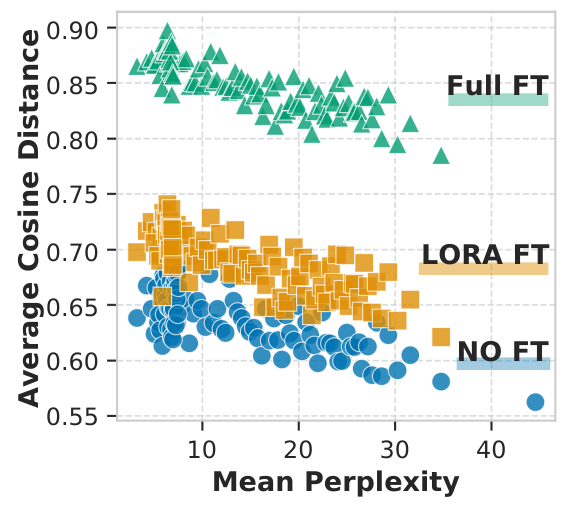}
        \caption{Effect of fine-Tuning on embedding separation.}
        \label{fig:finetune-sep}
    \end{minipage}
    \hfill  
    \begin{minipage}{0.67\textwidth}
        \centering
        \includegraphics[width=\linewidth]{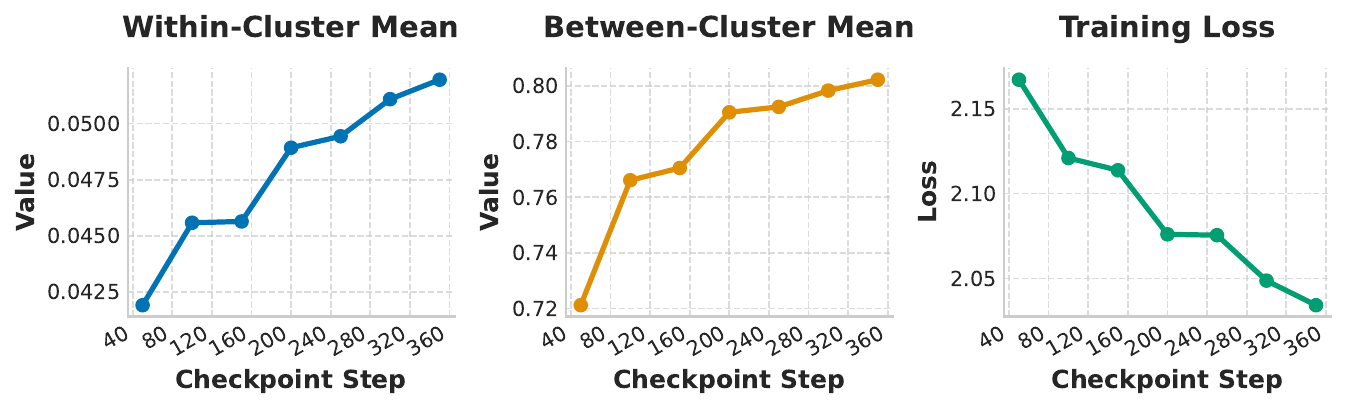}
        \caption{\textbf{Clustering metrics and training loss across training.}
    We form 500 clusters of WikiText‐103 contexts (each cluster contains 100
    semantically similar contexts that share the same 10‐gram continuation).
    During model training, both the \emph{within‐cluster} distance (left) and
    \emph{between‐cluster} distance (middle) consistently grow, while training
    loss (right) falls.}
    \label{fig:clustering-experiment}
    \end{minipage}
\end{figure}

\paragraph{(3) Fine-Tuning Effects.}
Finally, we examine how \emph{fine-tuning} reshapes the embedding space. We select the same model (LLaMA-3.2-1B) and apply either parameter-efficient (\emph{LoRA} \citep{DBLP:journals/corr/abs-2106-09685}) or \emph{full-parameter} fine-tuning on WikiText-103. Compared to the pre-trained model, both approaches \emph{increase} the average embedding separation on WikiText-103 (\Cref{fig:finetune-sep}). Full fine-tuning exerts a stronger effect, pushing text samples further apart overall; LoRA, which adapts only low-rank modules, effects smaller but still notable changes. The choice of fine-tuning thus influences how discriminative (i.e.\ “spread out”) the embeddings become in a specialized domain.

\subsection{Dispersion Within Semantic Clusters: A Finer-Grained Analysis}
\label{subsec:semantic-dispersion}

A natural question is whether better models merely push \emph{semantically dissimilar} contexts farther apart, or also increase distances among contexts that are \emph{semantically close}.  If the latter did not happen,
then we could imagine a scenario where highly related examples remain tightly clustered, but unrelated examples spread out, thereby inflating overall average distances.  We therefore directly measure dispersion among \emph{clusters} of highly similar contexts to see whether their internal geometry also expands over training.

\paragraph{Setup \& Metrics.}
We collect 10‐grams from the WikiText‐103 training set that appear at least 100 times.  For each such 10‐gram, we retrieve 100 occurrences, each accompanied by a 100‐token left context, and treat these contexts as a semantically similar \emph{cluster}. (Manual inspection confirmed that these context sets do indeed relate to the same event, entity, or theme.)  We construct a total of 500 such clusters.  We then track the contextual embeddings produced by a \(\textsc{LLaMA-3.2-1B}\) model at various checkpoints during training. Within each cluster, we compute the \emph{within‐cluster distance} as the average (cosine) distance from each embedding to the centroid of that cluster. Likewise, we define the \emph{between‐cluster distance} as the average (cosine) distance among the \emph{centroids} of all clusters.  Thus, the latter measures how far clusters are from each other in the embedding space, while the former measures how tightly each cluster is packed internally.

\paragraph{Results.}
\Cref{fig:clustering-experiment} shows that, as training proceeds and the loss decreases, both the average \emph{within‐cluster} distance and \emph{between‐cluster} distance increase.  Thus, the trend toward higher overall dispersion is not driven solely by pushing apart highly dissimilar contexts.  Even very similar contexts---which all share the same 10‐gram continuation---become more spread out in the latent space as the model learns, which indicates the model can effectively distinguish them despite the similarities. This in‐cluster expansion reinforces our hypothesis that better‐performing models tend to produce broader embedding geometries in \emph{all} regions of the latent space, not just pushing away dissimilar examples.

\section{Applications}

\subsection{Ranking Example Hardness without Labeled Data}
\label{sec:pred-downstream}

In many real-world scenarios, one receives a large \emph{unlabeled} query set and must decide, \emph{before} committing annotation or compute resources,  
(1) whether an off-the-shelf model will be sufficiently accurate and  
(2) which specific examples it is likely to get wrong.  
A reliable \emph{label-free} indicator would enable rapid model validation, automatic justification of \emph{easy} versus \emph{hard} instances, and targeted continued pre‑training on precisely those queries that the model currently struggles with.  
Because higher representation dispersion tracks lower perplexity (\S\ref{sec:analysis}), we hypothesize that dispersion can serve as such an indicator: if high dispersion coincides with correct predictions, then simply measuring distances allows us to rank examples by expected difficulty and flag likely errors without ever looking at ground‑truth labels.

\paragraph{Experimental protocol.}
To test this idea, we design a controlled experiment that varies the
\emph{fraction of correct answers} while keeping the input distribution fixed: (1) Collect a pool of question-answer pairs for a given dataset-model pair. (2) Partition the pool into \emph{correct} and \emph{incorrect} subsets by comparing the model’s answer with the ground truth. (3) For each desired accuracy level (0\%--100\% in 10\% increments), sample 100 queries that contain the requisite mix of correct and incorrect cases.(4) Extract the final-layer embeddings of these queries and compute their mean pairwise cosine distance, averaging over 10 random seeds.

\begin{figure}[t]
    \centering
    \includegraphics[width=0.75\linewidth]{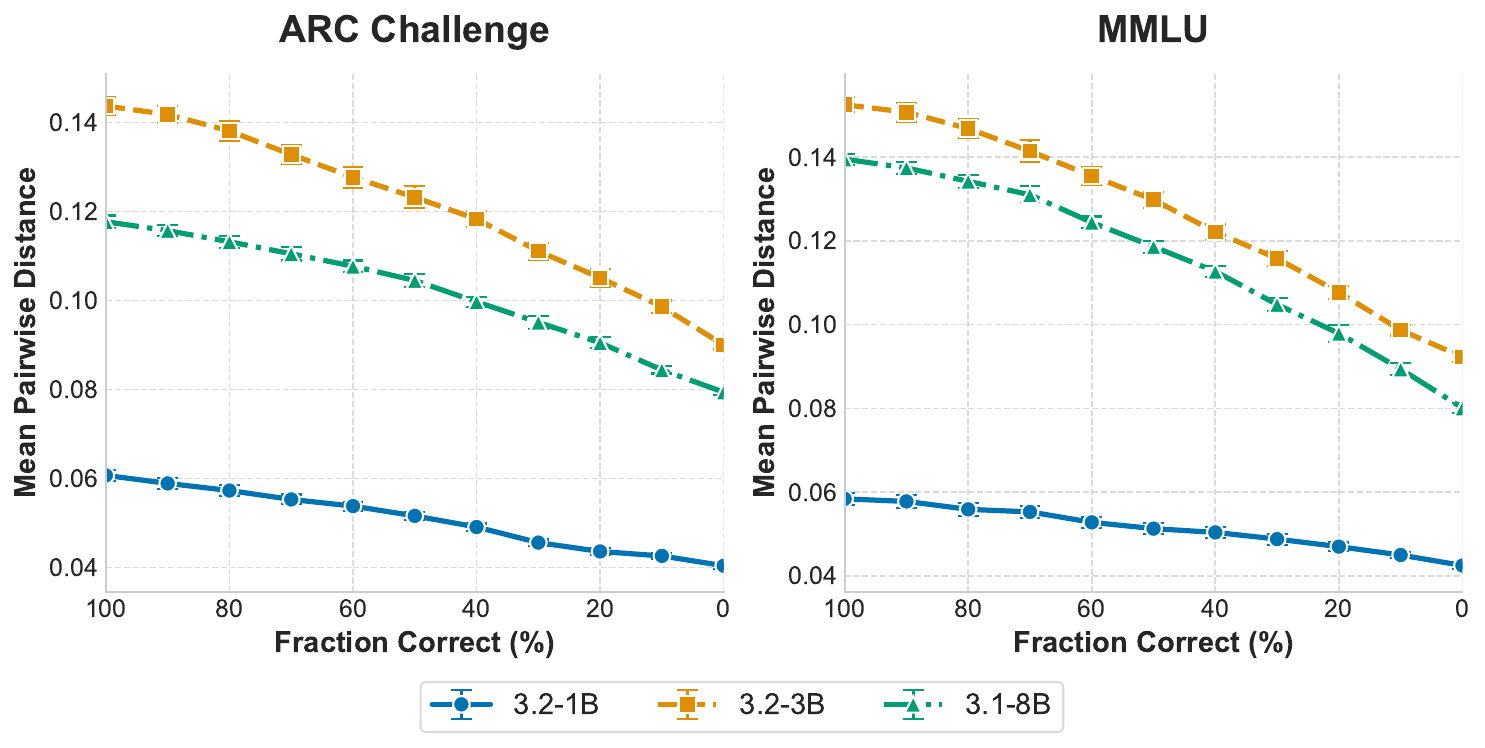}
    \caption{\textbf{Embedding distance vs.\ fraction correct.}
    Each point corresponds to a fixed mixture of correct/incorrect model
    predictions (x‐axis).  
    Error bars denote standard error over 10 random seeds.  
    Higher accuracy consistently aligns with larger mean pairwise distance. Similar trends are found in multilingual MMLU, HellaSwag, IFEval, and QuAIL. Details in \Cref{app:acc_additional_res} and \Cref{app:token_disp_new}.}
    \label{fig:fraction_correct_controlled}
\end{figure}

\paragraph{Results.}
\Cref{fig:fraction_correct_controlled} plots mean pairwise distance against the
fraction of correct predictions for several LLaMA variants on
\textsc{ARC‐Challenge} and \textsc{MMLU}.  
Across all models and datasets, \emph{accuracy rises monotonically with
dispersion}: slices that the model answers correctly exhibit markedly broader
geometry than those it answers incorrectly.  
Practically, a practitioner could therefore \emph{sort} an unlabeled dataset by
dispersion, inspect only the low‐dispersion tail to uncover failure modes, or
focus continued training on those ``hard'' queries. Taken together, these findings establish representation dispersion as a powerful, zero‐label indicator of \emph{relative hardness} and a principled tool for slice discovery and targeted data augmentation, rather than a fully calibrated predictor of absolute accuracy.

\subsection{Representation Dispersion for Model Selection}
\label{sec:embedding-dispersion-modelsel}

In practice, researchers and practitioners are often faced with choosing among numerous model variants—ranging from different instruction-tuned checkpoints to parameter-efficient adaptations or distilled models—while only having limited labeled data in the target domain. Exhaustively evaluating every checkpoint is typically prohibitively expensive, both in terms of computation and annotation resources. The tight link between representation dispersion and predictive accuracy established in \S\ref{sec:analysis} offers an attractive alternative: by simply measuring how broadly a model separates key tokens in its embedding space, one can obtain a geometric score that rapidly screens checkpoints \emph{within a fixed model family and tokenizer}.

\paragraph{Setup.} 

To operationalize this idea, we focus on task-relevant tokens that carry significant signal in the domain of interest, such as digits for mathematical reasoning, Python keywords for code generation, or legal terms for contract analysis. Let $\mathcal{T}\subset V$ denote a small set of such domain-specific tokens, and let $\overline{\mathcal{T}}$ denote a reference set of common, everyday language tokens. We use the model's original output token embeddings—that is, the rows of the output projection matrix—as provided after loading the model weights. This requires no forward passes or input data; all computations are performed directly on these pre-trained embeddings. We compute average pairwise distances—either cosine or Euclidean—among the embeddings of tokens within $\mathcal{T}$, within $\overline{\mathcal{T}}$, and between $\mathcal{T}$ and $\overline{\mathcal{T}}$.

Motivated by our empirical findings, we propose a single ``dispersion gap'' metric that succinctly captures both the distinctiveness of the domain-relevant tokens and their separation from generic language. Specifically, we define
\[
    \mathcal{G}=\text{within}(\mathcal{T})+\text{between}\bigl(\mathcal{T},\overline{\mathcal{T}}\bigr),
\]
where $\text{within}(\mathcal{T})$ denotes the mean pairwise distance among the domain tokens, and $\text{between}\bigl(\mathcal{T},\overline{\mathcal{T}}\bigr)$ is the mean distance between domain and reference tokens. Larger values of $\mathcal{G}$ indicate that the model both differentiates between domain-critical tokens and separates them from everyday vocabulary—a geometric pattern that, as Tables~\ref{tab:qwen-euclidean-dispersion}–\ref{tab:llama-code-cosine-dispersion} show, strongly correlates with higher task accuracy.

This approach is computationally efficient and entirely label-free: evaluating $\mathcal{G}$ requires only reading the model’s output embedding matrix and performing basic matrix operations on CPU, without any forward passes or GPU computation. In our experiments, ranking models by their dispersion gap within a given family and parameter scale consistently elevates the best-performing models in domains such as math and code, yielding gains of up to 40 accuracy points over clearly under-adapted variants. We therefore view the dispersion gap as a coarse but effective pre-filter: it reliably flags models that have not yet learned the relevant domain geometry, after which more expensive task-specific evaluation can be used to break ties among the remaining strong candidates.

\begin{figure}[h]
    \centering
    \includegraphics[width=0.8\textwidth]{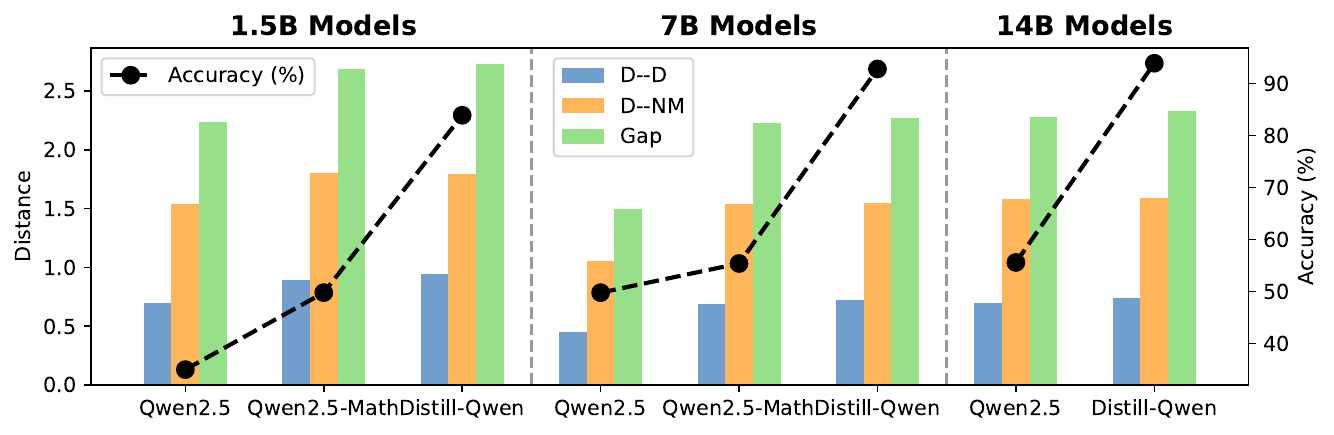}
    \hfill
    \includegraphics[width=0.51\textwidth]{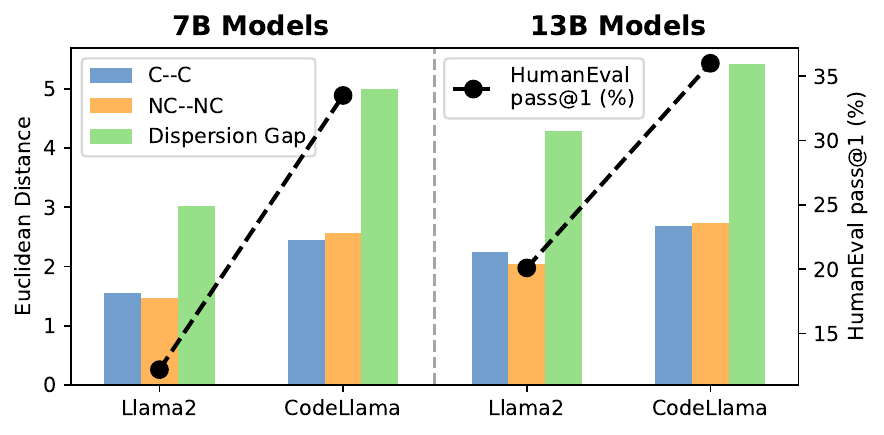}
    \caption{\textbf{Upper: Qwen on \textsc{MATH}.} Blue bars show $D$--$D$ (digit–digit) distances, orange bars show $D$--$NM$ (digit–non-math) distances, and green bars show their sum—the \emph{Dispersion Gap}. The black dashed line reports task accuracy (\%). \textbf{Lower: Llama/CodeLlama on \textsc{HumanEval}.}  Blue bars show $C$--$C$ (code-keyword) distances, orange bars show $NC$--$NC$ (non-code) distances, and green bars again give the \emph{Dispersion Gap}.  The black dashed line reports \textsc{pass@1} (\%). Bar heights therefore convey the three dispersion statistics, while the line traces model performance, mirroring the figure legends.  A full table of the underlying values is provided in \Cref{app:dispersion-performance}.}
    \label{fig:math_code_dispersion}
\end{figure}

\paragraph{Results.} \Cref{fig:math_code_dispersion} clearly demonstrates how the dispersion gap $\mathcal{G}$ aligns with downstream performance.  
In the \emph{upper} panel (Qwen on \textsc{MATH}), accuracy rises monotonically with the green ``Gap’’ bars both \emph{within} each parameter scale (e.g.\ 1.5 B\,\texttt{$\rightarrow$}\,7 B\,\texttt{$\rightarrow$}\,14 B) and \emph{across} fine-tuned variants (e.g.\ \textsc{Qwen2.5}, \textsc{Qwen2.5-Math}, \textsc{Distill-Qwen}).  
Spearman correlations between $\mathcal{G}$ and accuracy exceed $0.95$, and the dispersion gap perfectly ranks all nine checkpoints without a single mis-ordering. The \emph{lower} panel (Llama/CodeLlama on \textsc{HumanEval}) shows the same pattern:  
\textsc{CodeLlama} consistently exhibits both a larger gap and a higher {\small\textsc{pass@1}} rate than its \textsc{Llama2} counterpart at \emph{both} 7 B and 13 B scales, while the increase from 7 B to 13 B again boosts both metrics in lock-step. 
To confirm this pattern is robust beyond static model comparisons, \Cref{app:trajectory-analysis} extends our analysis to the training trajectory of Olmo-7B, showing that dispersion tracks performance improvements across 30 intermediate checkpoints with correlations exceeding $0.90$.

\subsection{Layer Selection for kNN-LM}
\label{sec:layer-selection}

Many retrieval-augmented language models build a datastore by using one of the sublayers in the final transformer block as the vector key \citep{khandelwal20generalization, he-etal-2021-efficient,  alon2022neurosymbolic, li2024chunkdistilledlanguagemodeling}. A central question in these methods is which internal representation of the transformer to use as the ``datastore key.'' In this section, we focus on \(k\)NN-LM, which augments a standard language model with a key-value memory of training examples. Specifically, at inference time, it retrieves the nearest neighbors of the current hidden state from that memory and interpolates their next-token distribution with the base LM’s distribution, often achieving lower perplexity on rare or out-of-distribution tokens. Recent evidence \citep{xu2023nearestneighborlanguagemodels} suggests that taking the attention-layer output (instead of the feed-forward layer output) often improves \(k\)NN-LM perplexity, but pinpointing the best layer can require expensive end-to-end trials. We show that measuring how widely each layer ``disperses'' its text embeddings (via average pairwise cosine distance) provides a lightweight, \emph{unsupervised} way to identify a layer likely to yield strong \(k\)NN-LM performance—without running the full interpolation pipeline.

\paragraph{Background.}
Consider the standard Transformer block as used by GPT-2. Let $ \mathbf{h}^{(l-1)} \in \mathbb{R}^{n \times d} $ denote the hidden states from block $l-1$. In block~$l$, the hidden states first pass through a multi-headed attention (MHA) sub-layer and residual connection:
$ \widehat{\mathbf{h}}^{(l)} = \mathbf{h}^{(l-1)} + \text{Dropout}\!\Bigl(\text{MHA}\bigl(\text{LayerNorm}(\mathbf{h}^{(l-1)})\bigr)\Bigr). $
We then apply another residual connection around a position-wise feed-forward network (FFN):
$ \mathbf{h}^{(l)} = \widehat{\mathbf{h}}^{(l)} + \text{Dropout}\!\Bigl(\text{FFN}\!\bigl(\text{LayerNorm}(\widehat{\mathbf{h}}^{(l)})\bigr)\Bigr). $
Thus, each block produces two intermediate sub-layer outputs:
$ \mathbf{h}^{(l)}_{\text{att}} = \widehat{\mathbf{h}}^{(l)} $
and
$ \mathbf{h}^{(l)}_{\text{ffn}} = \mathbf{h}^{(l)}. $
Following \citet{xu2023nearestneighborlanguagemodels}, we compare using
$ \mathbf{h}^{(L)}_{\text{att}} $
versus
$ \mathbf{h}^{(L)}_{\text{ffn}} $
from the \emph{final} block $L$ as the contextual key for $k$NN-LM.

\paragraph{Experimental Setup.}
We consider four members of the GPT-2 family—\textsc{distilgpt2} (82 M parameters), \textsc{gpt2-small} (117 M), \textsc{gpt2-medium} (345 M), and \textsc{gpt2-large} (774 M)—all trained with the same tokenizer and a context length of 1 024.  
For each checkpoint we draw \(N\in\{10,50,100\}\) non-overlapping 512-token \emph{chunks} randomly from the \textsc{Wikitext-103} validation split. Sampling is repeated 10 times with different random seeds, and the identical chunk indices are fed to both sub-layers so that any dispersion difference cannot be attributed to input variance. For every chunk we extract the \emph{last-token} hidden state produced by the final Transformer block’s  
(i) attention sub-layer output \(\mathbf h^{(L)}_{\text{att}}\) and  
(ii) feed-forward sub-layer output \(\mathbf h^{(L)}_{\text{ffn}}\).  
Given the resulting set of \(N\) vectors \(\{\mathbf h_i\}_{i=1}^{N}\), we measure their \emph{representation dispersion} as the average pairwise cosine distance.
We report the mean and standard deviation of dispersion across the 10 random repeats for every \(\langle\)model, \(N\), sub-layer\(\rangle\) triple.

\paragraph{Results.}
Table~\ref{tab:dispersion-layer-choice} shows two practitioner-relevant patterns.  
First, the hidden states taken \emph{after} the attention sub-layer are always more widely spread than those taken after the feed-forward sub-layer (e.g., 0.8045 vs.\ 0.6831 for \textsc{gpt2-large}, 0.6593 vs.\ 0.1865 for \textsc{gpt2-medium}), making \(\mathbf h^{(L)}_{\text{att}}\) the natural choice of key for \(k\)NN-LM.  
Second, dispersion can be estimated with striking efficiency: moving from 10 to 50 or 100 input chunks alters the mean by at most 1.5 \% and never changes the layer ranking, so profiling a model requires only about \(5{,}000\) tokens and a few milliseconds.  

\begin{table}[t]
\centering
\footnotesize
\begin{tabular}{lcccccc}
\toprule
& \multicolumn{2}{c}{\textbf{N=10}} & \multicolumn{2}{c}{\textbf{N=50}} & \multicolumn{2}{c}{\textbf{N=100}} \\
\cmidrule(lr){2-3} \cmidrule(lr){4-5} \cmidrule(lr){6-7}
\textbf{Model} & Attn & FFN & Attn & FFN & Attn & FFN \\
\midrule
\textsc{distilgpt2} & 0.83$\pm$0.02 & 0.33$\pm$0.05 & 0.83$\pm$0.01 & 0.30$\pm$0.02 & 0.82$\pm$0.01 & 0.30$\pm$0.01 \\
\textsc{gpt2-small} & 0.83$\pm$0.02 & 0.24$\pm$0.06 & 0.83$\pm$0.00 & 0.24$\pm$0.03 & 0.83$\pm$0.00 & 0.24$\pm$0.01 \\
\textsc{gpt2-medium} & 0.66$\pm$0.03 & 0.19$\pm$0.02 & 0.67$\pm$0.01 & 0.21$\pm$0.02 & 0.67$\pm$0.01 & 0.21$\pm$0.01 \\
\textsc{gpt2-large} & 0.80$\pm$0.04 & 0.68$\pm$0.06 & 0.79$\pm$0.01 & 0.66$\pm$0.02 & 0.80$\pm$0.01 & 0.68$\pm$0.03 \\
\bottomrule
\end{tabular}
\caption{Representation dispersion (average pairwise cosine distance) of the final block’s attention and feed-forward sub-layers for varying numbers of randomly sampled chunks \(N\). Higher values indicate more widely separated contextual embeddings.}
\label{tab:dispersion-layer-choice}
\end{table}

\subsection{Incorporating Representation Divergence into Training}
\label{sec:rd-regularization}

While typical cross-entropy training focuses purely on next-token prediction, several studies suggest that explicitly encouraging embedding separation can 
improve generalization and robustness in language modeling \citep{gunel2021supervisedcontrastivelearningpretrained, jain-etal-2023-contraclm}. Inspired by these findings, we augment the standard language-modeling loss with an auxiliary objective that \emph{pushes apart} hidden-state vectors, aiming to produce more discriminative representations.

\paragraph{Setup.}
We consider two scenarios: a \emph{single-domain} setting, where we train GPT-2 small on WikiText, and a \emph{cross-domain} setting, where we train on WikiText plus Python code. In both cases, let $\{\mathbf{h}_i\}_{i=1}^B$ be the final-layer hidden-state vectors for all token sequences in a batch (flattened across batch and sequence), where $B$ is the batch size. We normalize each vector to unit length, 
$\tilde{\mathbf{h}}_i = \mathbf{h}_i / \|\mathbf{h}_i\|$. To encourage wider separation, we compute an average pairwise cosine distance $d$ and then add an auxiliary loss $-\!d$ to the standard cross-entropy loss, weighted by a hyperparameter $\lambda$. 

In the \textbf{single-domain} setting, $d_{\text{avg}}$ is defined over all pairs in the same batch:
$d_{\text{avg}} = \frac{1}{B(B-1)} \sum_{i \neq j} \Bigl[ 1 - \tilde{\mathbf{h}}_i \cdot \tilde{\mathbf{h}}_j \Bigr]$.
In the \textbf{cross-domain} setting, we instead compute $d$ only across pairs drawn from different domains (Wiki vs.\ code) to push embeddings from each domain further apart:
$d = \frac{1}{|A||B|} \sum_{i\in A} \sum_{j\in B} \Bigl[ 1 - \tilde{\mathbf{h}}_i^{(\mathrm{A})} \cdot \tilde{\mathbf{h}}_j^{(\mathrm{B})} \Bigr]$.
The total loss in both settings is
$\mathcal{L}_{\mathrm{total}} = \mathcal{L}_{\mathrm{CE}} + \lambda \,\mathcal{L}_{\mathrm{aux}}$,
where $\mathcal{L}_{\mathrm{CE}}$ is the standard next-token cross-entropy and $\mathcal{L}_{\mathrm{aux}} = -d_{\text{avg}}$ (single-domain) or $-d$ (cross-domain), and $\lambda$ controls the strength of the auxiliary “spread-out” loss.. We clarify the relation between this auxiliary term and prior contrastive / repulsive losses in \Cref{app:contrastive}.

\paragraph{Results.}
Table~\ref{tab:aux_all_combined} reports test perplexities for various learning rates and auxiliary loss weights $\lambda$. In the \textbf{single-domain} (WikiText) setting, introducing the auxiliary spread-out loss yields a slight decrease in perplexity—typically 1–4 points—relative to the baseline, especially early in training. In the \textbf{cross-domain} (WikiText+Code) setting, the auxiliary loss produces a much more pronounced reduction in perplexity for both domains. For all learning rates, models trained with a moderate value of $\lambda$ achieve notably lower perplexity on both WikiText and code, suggesting that explicitly pushing apart representations from distinct domains leads to more specialized and di    scriminative features. This demonstrates that our approach is particularly effective when bridging heterogeneous data sources.

\begin{table}[h!]
\centering
\footnotesize
\begin{tabular}{c|ccccc|ccccc}
\toprule
\multirow{2}{*}{\textbf{LR}} 
  & \multicolumn{5}{c|}{\textbf{(a) Single-Domain (WikiText)}} 
  & \multicolumn{5}{c}{\textbf{(b) Cross-Domain (WikiText+Code)}} \\
\cmidrule(lr){2-6} \cmidrule(lr){7-11}
 & \boldmath$\lambda$ 
 & \multicolumn{2}{c}{\textbf{Step = 500}} 
 & \multicolumn{2}{c|}{\textbf{Step = 1000}} 
 & \boldmath$\lambda$ 
 & \multicolumn{2}{c}{\textbf{Step = 500}} 
 & \multicolumn{2}{c}{\textbf{Step = 1000}} \\
 & 
 & \textbf{Base} & \textbf{+Aux} 
 & \textbf{Base} & \textbf{+Aux} 
 & 
 & \textbf{Wiki} & \textbf{Code} 
 & \textbf{Wiki} & \textbf{Code} \\
\midrule
\multirow{2}{*}{$10^{-3}$} 
 & 0.0  & 226.1 & --    & 111.3 & --    
 & 0.0   & 295.6 & 36.9 & 171.7 & 23.8 \\
 & 0.1  & --    & 217.5 & --    & 108.2 
 & 0.001 & 270.8 & 34.5 & 158.8 & 22.3 \\
\midrule
\multirow{2}{*}{$7\times10^{-4}$}
 & 0.0  & 195.0 & --    & 96.7  & --    
 & 0.0   & 304.4 & 35.4 & 175.7 & 22.8 \\
 & 0.1  & --    & 193.8 & --    & 93.6  
 & 0.01  & 255.2 & 31.9 & 150.2 & 20.8 \\
\midrule
\multirow{2}{*}{$5\times10^{-4}$}
 & 0.0  & 166.2 & --    & 83.0  & --    
 & 0.0   & 268.5 & 33.7 & 166.9 & 22.1 \\
 & 0.1  & --    & 165.6 & --    & 82.0  
 & 0.02  & 253.3 & 30.5 & 155.2 & 20.5 \\
\bottomrule
\end{tabular}
\caption{\textbf{Auxiliary spread-out loss improves perplexity in both single- and cross-domain settings.} 
Left: single-domain (WikiText) results; right: cross-domain (WikiText+Code) results. $\lambda$ is the auxiliary loss weight, chosen by validation for each learning rate. We report test-set perplexities at 500 and 1000 steps.}
\label{tab:aux_all_combined}
\end{table}

\section{Related Work}
\label{sec:related_work}

\paragraph{Geometric Analysis of Embeddings in Language Models.}
A growing body of work has examined the geometry of hidden representations in large language models (LLMs). Early studies identified an anisotropy problem, where embeddings collapse into a narrow cone and lose expressiveness at deeper layers \citep{DBLP:conf/iclr/MuV18, DBLP:conf/emnlp/Ethayarajh19,DBLP:conf/iclr/GaoHTQWL19,DBLP:conf/emnlp/LiZHWYL20,noci2022signal}. 
Recent work uses intrinsic dimension (ID) estimators to trace how representation manifolds evolve across layers \citep{valeriani2023geometry}, linking geometry to performance through, for example, distinguishing human vs.\ machine text \citep{tulchinskii2023intrinsic}, predicting data compressibility \citep{cheng-etal-2023-bridging}, and revealing simplex-like structures for categorical concepts \citep{park2025geometrycategoricalhierarchicalconcepts}.

The study most closely related to ours is \citet{viswanathan2025the}, which also analyzes token-level embedding distributions and observes cosine similarity rises when tokens in the prompt are shuffled. However, their work remains largely descriptive. In contrast, we show how representation dispersion can predict and improve perplexity and downstream accuracy, making geometric insights actionable for model evaluation and selection.



\paragraph{Mechanistic Interpretability.}
Our work offers a geometric perspective on interpretability, complementing research into model mechanisms. This field often reverse-engineers models into interpretable circuits \citep{elhage2021mathematical, bereska2024mechanistic}, identifying specific components like induction heads for in-context learning \citep{olsson2022incontextlearninginductionheads} or analyzing training dynamics through layer-wise gradients \citep{li2025happenedllmslayerstrained, li2025instructionreasoningdatashape}. Rather than dissecting individual components, we provide a higher-level, component-agnostic metric by quantifying the dispersion of representations—an abstract but behaviorally meaningful property of the model’s internal geometry.

Our method also differs from post-hoc attribution and probing techniques. Attribution methods like Integrated Gradients assign credit to inputs for a given output \citep{sundararajan2017axiomatic, NIPS2017_8a20a862}, while probing trains auxiliary classifiers on hidden states to identify encoded linguistic features \citep{belinkov-2022-probing}. These approaches typically require labeled data or external probes. In contrast, representational dispersion is a label-free, intrinsic measure; we directly relate the geometry of hidden states to the model's own performance metrics, such as perplexity and downstream accuracy.

\section{Conclusion}
\label{sec:conclusion}


In this work, we showed that representation dispersion serves as both a practical diagnostic and training signal for language models. Moving forward, we aim to investigate how representation dispersion interacts with other design choices—such as architectural variations or tokenization strategies—and whether additional regularization signals might further strengthen model robustness and interpretability. We hope these directions will inspire new ways to harness embedding geometry for next‐generation language modeling and related tasks.


\newpage

\section*{Statement on LLM Usage}
We acknowledge the use of Large Language Models (LLMs) to assist in the preparation of this manuscript. Specifically, LLMs were utilized to improve grammar and clarity, aid in literature discovery, and generate boilerplate code snippets for our experiments and testing scripts. The authors have carefully reviewed and edited all LLM-generated outputs and take full responsibility for the final content and scientific integrity of this work.

\bibliography{references}

@misc{friedman2023vendiscorediversityevaluation,
      title={The Vendi Score: A Diversity Evaluation Metric for Machine Learning}, 
      author={Dan Friedman and Adji Bousso Dieng},
      year={2023},
      eprint={2210.02410},
      archivePrefix={arXiv},
      primaryClass={cs.LG},
      url={https://arxiv.org/abs/2210.02410}, 
}

@inproceedings{xu2023nearestneighborlanguagemodels,
author = {Xu, Frank F. and Alon, Uri and Neubig, Graham},
title = {Why do nearest neighbor language models work?},
year = {2023},
publisher = {JMLR.org},
booktitle = {Proceedings of the 40th International Conference on Machine Learning},
articleno = {1596},
numpages = {17},
location = {Honolulu, Hawaii, USA},
series = {ICML'23}
}

@inproceedings{khandelwal20generalization,
  title={{Generalization through Memorization: Nearest Neighbor Language Models}},
  author={Khandelwal, Urvashi and Levy, Omer and Jurafsky, Dan and Zettlemoyer, Luke and Lewis, Mike},
  booktitle={International Conference on Learning Representations (ICLR)},
  year={2020}
}

@article{elhage2021mathematical,
   title={A Mathematical Framework for Transformer Circuits},
   author={Elhage, Nelson and Nanda, Neel and Olsson, Catherine and Henighan, Tom and Joseph, Nicholas and Mann, Ben and Askell, Amanda and Bai, Yuntao and Chen, Anna and Conerly, Tom and DasSarma, Nova and Drain, Dawn and Ganguli, Deep and Hatfield-Dodds, Zac and Hernandez, Danny and Jones, Andy and Kernion, Jackson and Lovitt, Liane and Ndousse, Kamal and Amodei, Dario and Brown, Tom and Clark, Jack and Kaplan, Jared and McCandlish, Sam and Olah, Chris},
   year={2021},
   journal={Transformer Circuits Thread},
   note={https://transformer-circuits.pub/2021/framework/index.html}
}

@misc{li2024chunkdistilledlanguagemodeling,
      title={Chunk-Distilled Language Modeling}, 
      author={Yanhong Li and Karen Livescu and Jiawei Zhou},
      year={2024},
      eprint={2501.00343},
      archivePrefix={arXiv},
      primaryClass={cs.CL},
      url={https://arxiv.org/abs/2501.00343}, 
}

@inproceedings{he-etal-2021-efficient,
    title = "Efficient Nearest Neighbor Language Models",
    author = "He, Junxian  and
      Neubig, Graham  and
      Berg-Kirkpatrick, Taylor",
    editor = "Moens, Marie-Francine  and
      Huang, Xuanjing  and
      Specia, Lucia  and
      Yih, Scott Wen-tau",
    booktitle = "Proceedings of the 2021 Conference on Empirical Methods in Natural Language Processing",
    month = nov,
    year = "2021",
    address = "Online and Punta Cana, Dominican Republic",
    publisher = "Association for Computational Linguistics",
    url = "https://aclanthology.org/2021.emnlp-main.461",
    doi = "10.18653/v1/2021.emnlp-main.461",
    pages = "5703--5714",
}

@inproceedings{alon2022neurosymbolic, 
    title={Neuro-Symbolic Language Modeling with Automaton-augmented Retrieval},
    author={Alon, Uri and Xu, Frank and He, Junxian and Sengupta, Sudipta and Roth, Dan and Neubig, Graham},
    booktitle={	 {Proceedings of the 39th International Conference on Machine Learning}},
    pages={	 {468--485}},
    year={	 {2022}},
    editor={	 {Chaudhuri, Kamalika and Jegelka, Stefanie and Song, Le and Szepesvari, Csaba and Niu, Gang and Sabato, Sivan}},
    volume={	 {162}},
    series={	 {Proceedings of Machine Learning Research}},
    month={	 {17--23 Jul}},
    publisher={PMLR},
    url={	 {https://proceedings.mlr.press/v162/alon22a.html}},
}

@inproceedings{DBLP:conf/emnlp/Ethayarajh19,
  author    = {Kawin Ethayarajh},
  title     = {How Contextual are Contextualized Word Representations? Comparing
               the Geometry of BERT, ELMo, and {GPT-2} Embeddings},
  booktitle = {EMNLP-IJCNLP 2019, Hong Kong, China},
  pages     = {55--65},
  year      = {2019},
  bibsource = {dblp computer science bibliography, https://dblp.org}
}

@inproceedings{DBLP:conf/iclr/GaoHTQWL19,
  author    = {Jun Gao and
               Di He and
               Xu Tan and
               Tao Qin and
               Liwei Wang and
               Tie{-}Yan Liu},
  title     = {Representation Degeneration Problem in Training Natural Language Generation
               Models},
  booktitle = {7th International Conference on Learning Representations, {ICLR} 2019,
               New Orleans, LA, USA},
  year      = {2019},
  bibsource = {dblp computer science bibliography, https://dblp.org}
}

@inproceedings{DBLP:conf/emnlp/LiZHWYL20,
  author    = {Bohan Li and
               Hao Zhou and
               Junxian He and
               Mingxuan Wang and
               Yiming Yang and
               Lei Li},
  editor    = {Bonnie Webber and
               Trevor Cohn and
               Yulan He and
               Yang Liu},
  title     = {On the Sentence Embeddings from Pre-trained Language Models},
  booktitle = {Proceedings of the 2020 Conference on EMNLP 2020, Online, November 16-20, 2020},
  pages     = {9119--9130},
  year      = {2020},
  bibsource = {dblp computer science bibliography, https://dblp.org}
}

@inproceedings{DBLP:conf/iclr/MuV18,
  author    = {Jiaqi Mu and
               Pramod Viswanath},
  title     = {All-but-the-Top: Simple and Effective Postprocessing for Word Representations},
  booktitle = {{ICLR} 2018,
               Vancouver, BC, Canada},
  year      = {2018},
  bibsource = {dblp computer science bibliography, https://dblp.org}
}

@article{noci2022signal,
  title={Signal propagation in transformers: Theoretical perspectives and the role of rank collapse},
  author={Noci, Lorenzo and Anagnostidis, Sotiris and Biggio, Luca and Orvieto, Antonio and Singh, Sidak Pal and Lucchi, Aurelien},
  journal={Advances in Neural Information Processing Systems},
  volume={35},
  pages={27198--27211},
  year={2022}
}

@article{valeriani2023geometry,
  title={The geometry of hidden representations of large transformer models},
  author={Valeriani, Lucrezia and Doimo, Diego and Cuturello, Francesca and Laio, Alessandro and Ansuini, Alessio and Cazzaniga, Alberto},
  journal={Advances in Neural Information Processing Systems},
  volume={36},
  pages={51234--51252},
  year={2023}
}

@article{tulchinskii2023intrinsic,
  title={Intrinsic dimension estimation for robust detection of ai-generated texts},
  author={Tulchinskii, Eduard and Kuznetsov, Kristian and Kushnareva, Laida and Cherniavskii, Daniil and Nikolenko, Sergey and Burnaev, Evgeny and Barannikov, Serguei and Piontkovskaya, Irina},
  journal={Advances in Neural Information Processing Systems},
  volume={36},
  pages={39257--39276},
  year={2023}
}

@inproceedings{cheng-etal-2023-bridging,
    title = "Bridging Information-Theoretic and Geometric Compression in Language Models",
    author = "Cheng, Emily  and
      Kervadec, Corentin  and
      Baroni, Marco",
    editor = "Bouamor, Houda  and
      Pino, Juan  and
      Bali, Kalika",
    booktitle = "Proceedings of the 2023 Conference on Empirical Methods in Natural Language Processing",
    month = dec,
    year = "2023",
    address = "Singapore",
    publisher = "Association for Computational Linguistics",
    url = "https://aclanthology.org/2023.emnlp-main.762/",
    doi = "10.18653/v1/2023.emnlp-main.762",
    pages = "12397--12420"
}

@misc{park2025geometrycategoricalhierarchicalconcepts,
      title={The Geometry of Categorical and Hierarchical Concepts in Large Language Models}, 
      author={Kiho Park and Yo Joong Choe and Yibo Jiang and Victor Veitch},
      year={2025},
      eprint={2406.01506},
      archivePrefix={arXiv},
      primaryClass={cs.CL},
      url={https://arxiv.org/abs/2406.01506}, 
}

@misc{
viswanathan2025the,
title={The Geometry of Tokens in Internal Representations of Large Language Models},
author={Karthik Viswanathan and Yuri Gardinazzi and Giada Panerai and Alberto Cazzaniga and Matteo Biagetti},
year={2025},
url={https://openreview.net/forum?id=an3jH2qD2r}
}

@article{
bereska2024mechanistic,
title={Mechanistic Interpretability for {AI} Safety - A Review},
author={Leonard Bereska and Stratis Gavves},
journal={Transactions on Machine Learning Research},
issn={2835-8856},
year={2024},
url={https://openreview.net/forum?id=ePUVetPKu6},
note={Survey Certification, Expert Certification}
}

@misc{olsson2022incontextlearninginductionheads,
      title={In-context Learning and Induction Heads}, 
      author={Catherine Olsson and Nelson Elhage and Neel Nanda and Nicholas Joseph and Nova DasSarma and Tom Henighan and Ben Mann and Amanda Askell and Yuntao Bai and Anna Chen and Tom Conerly and Dawn Drain and Deep Ganguli and Zac Hatfield-Dodds and Danny Hernandez and Scott Johnston and Andy Jones and Jackson Kernion and Liane Lovitt and Kamal Ndousse and Dario Amodei and Tom Brown and Jack Clark and Jared Kaplan and Sam McCandlish and Chris Olah},
      year={2022},
      eprint={2209.11895},
      archivePrefix={arXiv},
      primaryClass={cs.LG},
      url={https://arxiv.org/abs/2209.11895}, 
}

@inproceedings{sundararajan2017axiomatic,
  title={Axiomatic attribution for deep networks},
  author={Sundararajan, Mukund and Taly, Ankur and Yan, Qiqi},
  booktitle={International conference on machine learning},
  pages={3319--3328},
  year={2017},
  organization={PMLR}
}

@inproceedings{NIPS2017_8a20a862,
 author = {Lundberg, Scott M and Lee, Su-In},
 booktitle = {Advances in Neural Information Processing Systems},
 editor = {I. Guyon and U. Von Luxburg and S. Bengio and H. Wallach and R. Fergus and S. Vishwanathan and R. Garnett},
 pages = {},
 publisher = {Curran Associates, Inc.},
 title = {A Unified Approach to Interpreting Model Predictions},
 url = {https://proceedings.neurips.cc/paper_files/paper/2017/file/8a20a8621978632d76c43dfd28b67767-Paper.pdf},
 volume = {30},
 year = {2017}
}

@article{belinkov-2022-probing,
    title = "Probing Classifiers: Promises, Shortcomings, and Advances",
    author = "Belinkov, Yonatan",
    journal = "Computational Linguistics",
    volume = "48",
    number = "1",
    month = mar,
    year = "2022",
    address = "Cambridge, MA",
    publisher = "MIT Press",
    url = "https://aclanthology.org/2022.cl-1.7/",
    doi = "10.1162/coli_a_00422",
    pages = "207--219"
}

@article{DBLP:journals/corr/abs-2106-09685,
  author       = {Edward J. Hu and
                  Yelong Shen and
                  Phillip Wallis and
                  Zeyuan Allen{-}Zhu and
                  Yuanzhi Li and
                  Shean Wang and
                  Weizhu Chen},
  title        = {LoRA: Low-Rank Adaptation of Large Language Models},
  journal      = {CoRR},
  volume       = {abs/2106.09685},
  year         = {2021},
  url          = {https://arxiv.org/abs/2106.09685},
  eprinttype    = {arXiv},
  eprint       = {2106.09685},
  bibsource    = {dblp computer science bibliography, https://dblp.org}
}

@misc{gunel2021supervisedcontrastivelearningpretrained,
      title={Supervised Contrastive Learning for Pre-trained Language Model Fine-tuning}, 
      author={Beliz Gunel and Jingfei Du and Alexis Conneau and Ves Stoyanov},
      year={2021},
      eprint={2011.01403},
      archivePrefix={arXiv},
      primaryClass={cs.CL},
      url={https://arxiv.org/abs/2011.01403}, 
}

@inproceedings{jain-etal-2023-contraclm,
    title = "{C}ontra{CLM}: Contrastive Learning For Causal Language Model",
    author = "Jain, Nihal  and
      Zhang, Dejiao  and
      Ahmad, Wasi Uddin  and
      Wang, Zijian  and
      Nan, Feng  and
      Li, Xiaopeng  and
      Tan, Ming  and
      Nallapati, Ramesh  and
      Ray, Baishakhi  and
      Bhatia, Parminder  and
      Ma, Xiaofei  and
      Xiang, Bing",
    editor = "Rogers, Anna  and
      Boyd-Graber, Jordan  and
      Okazaki, Naoaki",
    booktitle = "Proceedings of the 61st Annual Meeting of the Association for Computational Linguistics (Volume 1: Long Papers)",
    month = jul,
    year = "2023",
    address = "Toronto, Canada",
    publisher = "Association for Computational Linguistics",
    url = "https://aclanthology.org/2023.acl-long.355/",
    doi = "10.18653/v1/2023.acl-long.355",
    pages = "6436--6459"
}

@misc{li2025instructionreasoningdatashape,
      title={How Instruction and Reasoning Data shape Post-Training: Data Quality through the Lens of Layer-wise Gradients}, 
      author={Ming Li and Yanhong Li and Ziyue Li and Tianyi Zhou},
      year={2025},
      eprint={2504.10766},
      archivePrefix={arXiv},
      primaryClass={cs.LG},
      url={https://arxiv.org/abs/2504.10766}, 
}

@misc{li2025happenedllmslayerstrained,
      title={What Happened in LLMs Layers when Trained for Fast vs. Slow Thinking: A Gradient Perspective}, 
      author={Ming Li and Yanhong Li and Tianyi Zhou},
      year={2025},
      eprint={2410.23743},
      archivePrefix={arXiv},
      primaryClass={cs.CL},
      url={https://arxiv.org/abs/2410.23743}, 
}
\bibliographystyle{iclr2026_conference}
\clearpage

\appendix
\clearpage

\begin{center}
    \Large{\textbf{Technical Appendices}}
\end{center}

\setcounter{page}{11}
\startcontents[appendix]
\printcontents[appendix]{ }{0}{\section*{Organization of Contents}}

\newpage

\section{Supplemental Materials for Representation Geometry Analysis}
\label{app:rep-geometry}

\subsection{Details regarding Sequence-level Perplexity Experiments}

\subsubsection{Datasets and Models}
In this section, we provide additional experimental details and visualizations that supplement our main empirical analysis in \S\ref{sec:analysis}. We study a range of standard language modeling datasets, including the Salesforce/wikitext \footnote{\url{http://huggingface.co/datasets/Salesforce/wikitext}}, abisee/cnn\_dailymail \footnote{\url{https://huggingface.co/datasets/abisee/cnn_dailymail}}, and ccdv/pubmed-summarization \footnote{\url{https://huggingface.co/datasets/ccdv/pubmed-summarization}}, covering text segments in diverse domains. 

For the models, our experiments encompass:
\begin{itemize}
    \item \textbf{Llama} families: \texttt{meta-llama/Llama-3.2-1B}, \texttt{meta-llama/Llama-3.2-3B}, \texttt{meta-llama/Llama-3.1-8B}
    \item \textbf{Gemma} families: \texttt{google/gemma-2-2b}, \texttt{google/gemma-2-9b}
    \item \textbf{Mistral}: \texttt{mistralai/Mistral-7B-v0.1}
    \item \textbf{Phi}: \texttt{microsoft/phi-2}
    \item \textbf{Qwen} families: \texttt{Qwen/Qwen2.5-0.5B}, \texttt{Qwen/Qwen2.5-3B}, \texttt{Qwen/Qwen2.5-7B}
\end{itemize}

We use the Hugging Face implementation of the above models. All models are standard decoder-only Transformers, for which we collect final-layer embeddings on randomly selected text segments. In line with Equation~\ref{eq:avg-dist} of the main paper, we measure average pairwise cosine distance to quantify how ``spread out'' their representations are.

\subsubsection{Procedure for Mean-Perplexity vs.\ Dispersion Analysis}
\label{app:ppl-dispersion-detail}

Here, we outline the steps needed to produce a mean-perplexity vs.\ representation-dispersion plot:

\begin{enumerate}[%
  label=\textbf{Step~\arabic*:}, 
  leftmargin=60pt, 
  itemsep=0.75ex
]
  \item Randomly sample 100{,}000 segments (e.g., 512 tokens each) from the data.

  \item For each segment:
    \begin{enumerate}[%
      label=\alph*), 
      leftmargin=1.5em,
      itemsep=0.5ex
    ]
      \item Compute its perplexity over the full sequence.
      \item Record the final-layer hidden states for later analysis.
    \end{enumerate}

  \item Sort all segments by their computed perplexity.

  \item Group the sorted segments into bins (e.g., 100 segments per bin) and record each bin’s mean perplexity.

  \item Perform uniform sampling in perplexity space on these bins to ensure coverage of low-, mid-, and high-perplexity regions.

  \item For each uniformly sampled bin:
    \begin{enumerate}[%
      label=\alph*), 
      leftmargin=1.5em,
      itemsep=0.5ex
    ]
      \item Retrieve the saved hidden states.
      \item Calculate pairwise distances (e.g., average cosine distance) among the segment embeddings.
    \end{enumerate}

  \item Produce the final mapping of mean perplexity to average pairwise distance.
\end{enumerate}

\vspace{-1.5em}
\paragraph{Uniform Perplexity Sampling.}
Since random sampling of text segments often yields a distribution heavily concentrated around moderate perplexities, we use a uniform sampling scheme to cover both low- and high-perplexity ``tails.'' 
The pseudocode below highlights the procedure used in \textbf{Step 5} (Algorithm~\ref{alg:uniform-ppl}):

\begin{algorithm}[H]
  \caption{Uniform Perplexity Binning}
  \label{alg:uniform-ppl}
  \begin{algorithmic}[1]
    \Require A sorted list of $G$ perplexity bins $\{b_1,\dots,b_G\}$ (with means $m_1\le\cdots\le m_G$)
    \Require Desired number of bins $K$
    \Ensure A set of $K$ bins sampled uniformly in perplexity

    \State $m_{\min}\gets m_1$; \quad $m_{\max}\gets m_G$
    \State Define targets 
    \[
      t_k \;=\; m_{\min} + \frac{k-1}{K-1}\,(m_{\max}-m_{\min})
      \quad\text{for }k=1,\dots,K
    \]
    \State $\mathit{selected}\gets\emptyset$
    \For{$k\gets1$ \textbf{to} $K$}
      \State find $j$ s.t.\ $m_j$ is closest to $t_k$
      \State $\mathit{selected}\gets \mathit{selected}\cup\{j\}$
    \EndFor
    \If{$|\mathit{selected}|<K$}
      \State add extra bins from the sorted list until you have $K$
    \EndIf
    \State \Return $\{\,b_j : j\in\mathit{selected}\}$
  \end{algorithmic}
\end{algorithm}

This ensures we sample across the entire perplexity spectrum, capturing both rare, low-ppl segments and rare, high-ppl segments. 
With these selected bins in hand, we can then compute the final-layer embeddings and measure representation dispersion to obtain a mean-ppl vs.\ dispersion plot.

\bigskip

\subsubsection{Additional Visualizations}
\label{subsec:all_vis}
Below, we present the full set of perplexity-versus-dispersion plots referenced in \S\ref{subsec:global-observations}. 
For each dataset and model, we group 100{,}000 text segments into perplexity bins and compute their average pairwise representation distances. As described in the main text, we observe a negative correlation between sequence-level perplexity and representation dispersion.

\begin{figure}[h]
    \centering
    \includegraphics[width=0.65\textwidth]{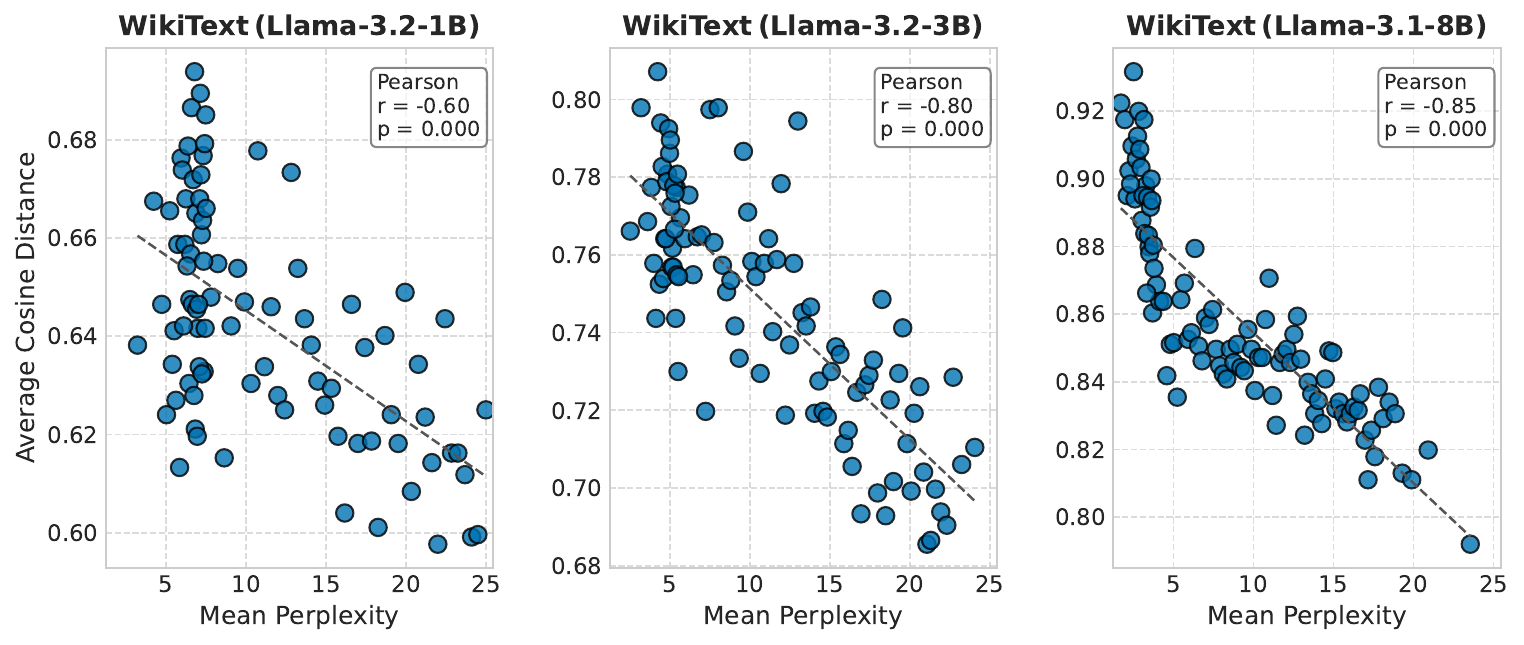}
    \caption{Perplexity vs.\ Average Pairwise Cosine Distance on Wikitext-103 (Llama family).}
    \label{fig:appendix-wt103-llama}
\end{figure}

\begin{figure}[h]
    \centering
    \includegraphics[width=0.45\textwidth]{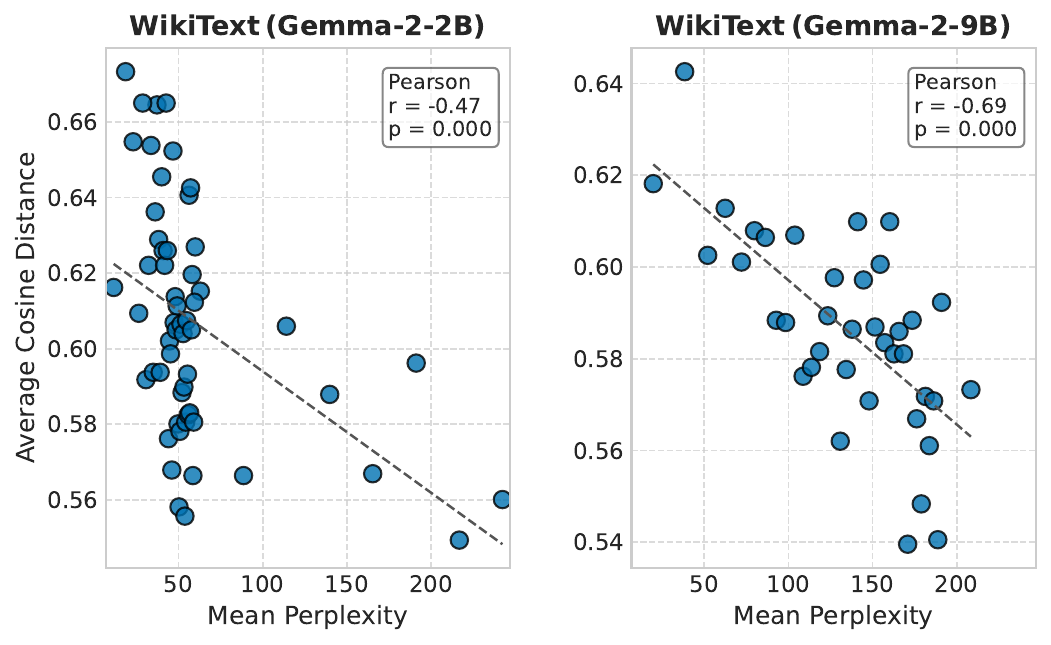}
    \caption{Perplexity vs.\ Average Pairwise Cosine Distance on Wikitext-103 (Gemma family).}
    \label{fig:appendix-wt103-gemma}
\end{figure}

\begin{figure}[h]
    \centering
    \includegraphics[width=0.65\textwidth]{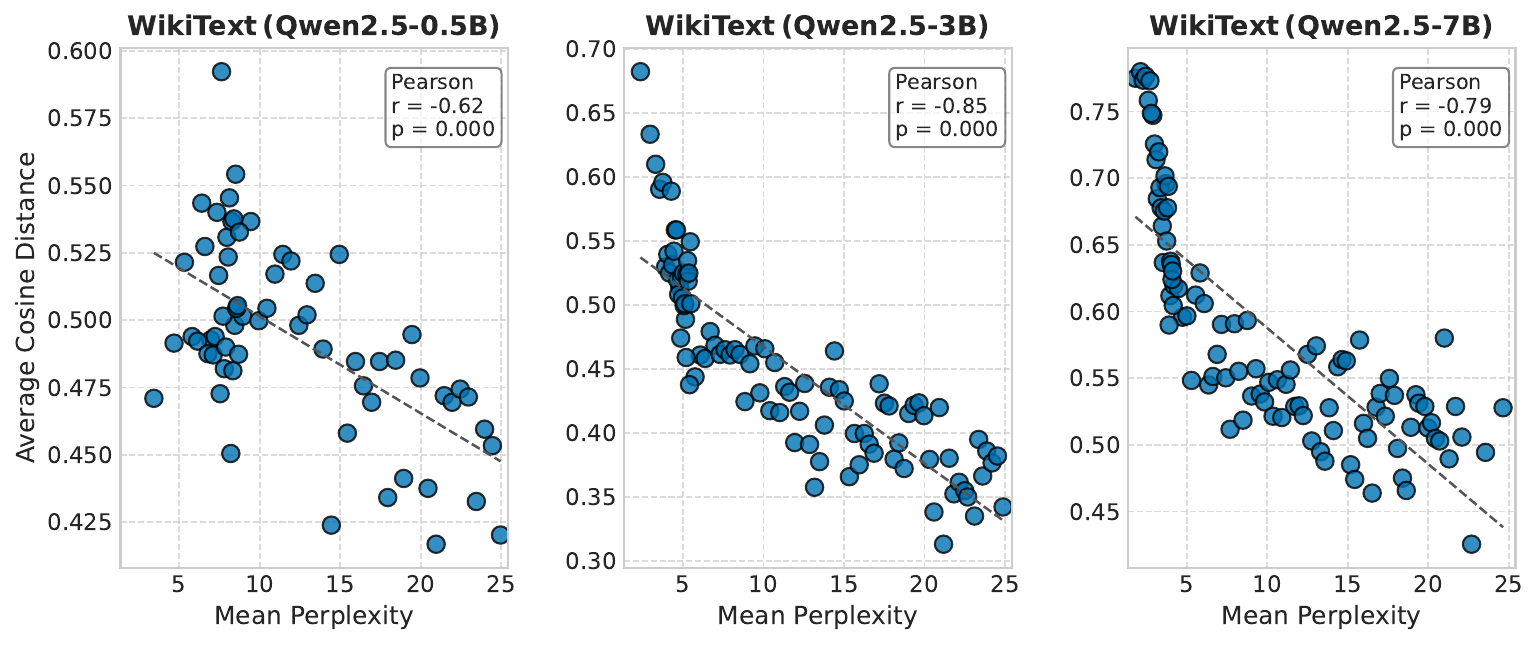}
    \caption{Perplexity vs.\ Average Pairwise Cosine Distance on Wikitext-103 (Qwen family).}
    \label{fig:appendix-wt103-qwen}
\end{figure}

\begin{figure}[h]
    \centering
    \includegraphics[width=0.45\textwidth]{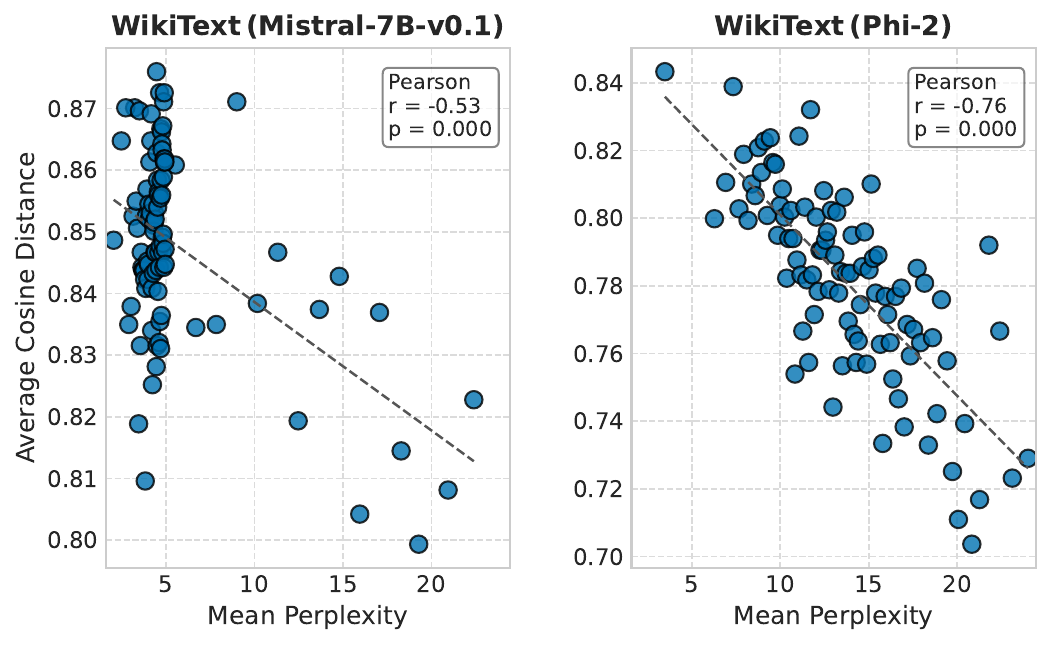}
    \caption{Perplexity vs.\ Average Pairwise Cosine Distance on Wikitext-103 (Mistral) and Phi.}
    \label{fig:appendix-wt103-phi-mistral}
\end{figure}

\begin{figure}[h]
    \centering
    \includegraphics[width=0.65\textwidth]{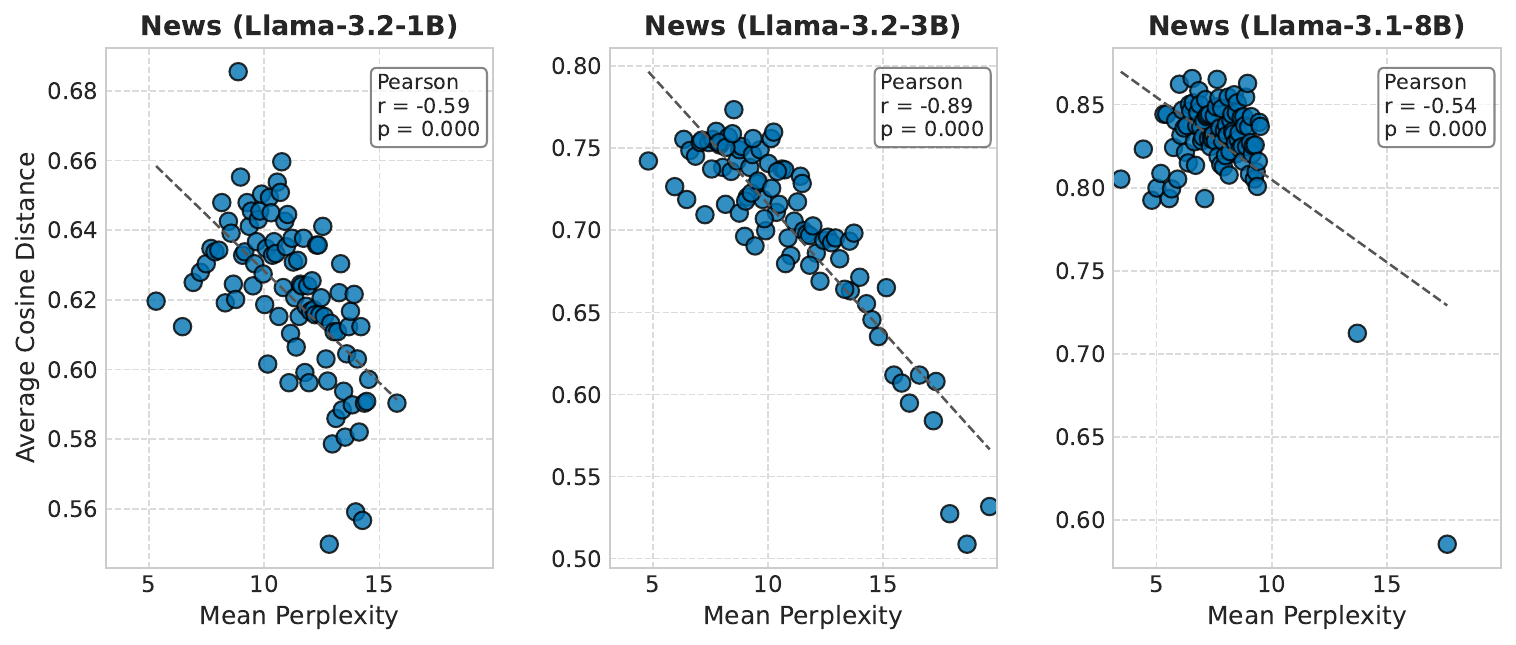}
    \caption{Perplexity vs.\ Average Pairwise Cosine Distance on CNN DailyMail (Llama family).}
    \label{fig:appendix-cnn-llama}
\end{figure}

\begin{figure}[h]
    \centering
    \includegraphics[width=0.45\textwidth]{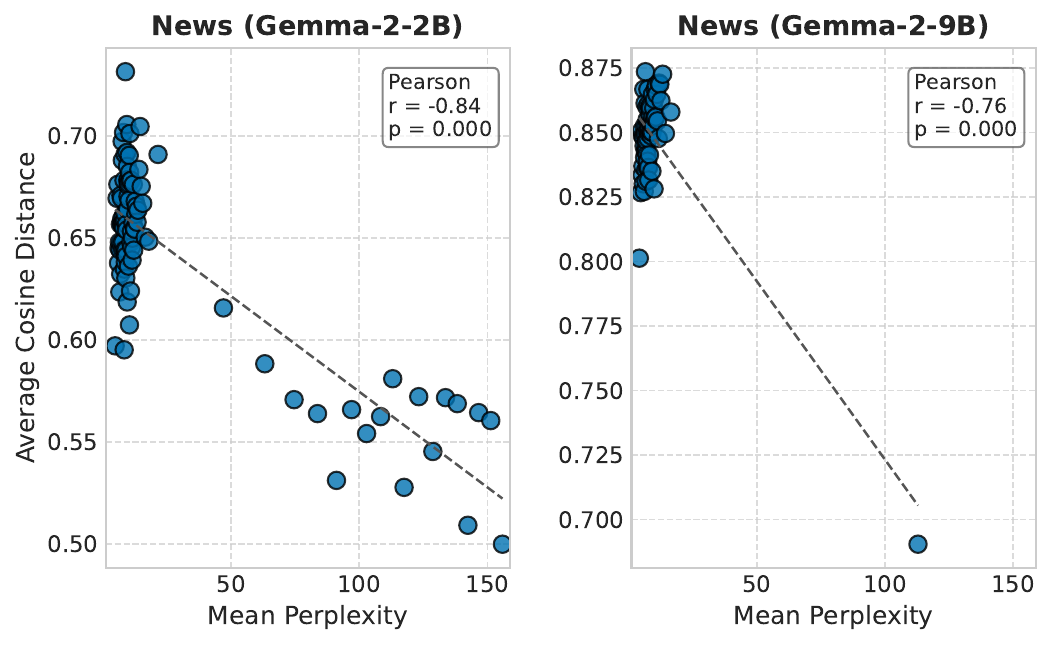}
    \caption{Perplexity vs.\ Average Pairwise Cosine Distance on CNN DailyMail (Gemma family).}
    \label{fig:appendix-cnn-gemma}
\end{figure}

\begin{figure}[h]
    \centering
    \includegraphics[width=0.65\textwidth]{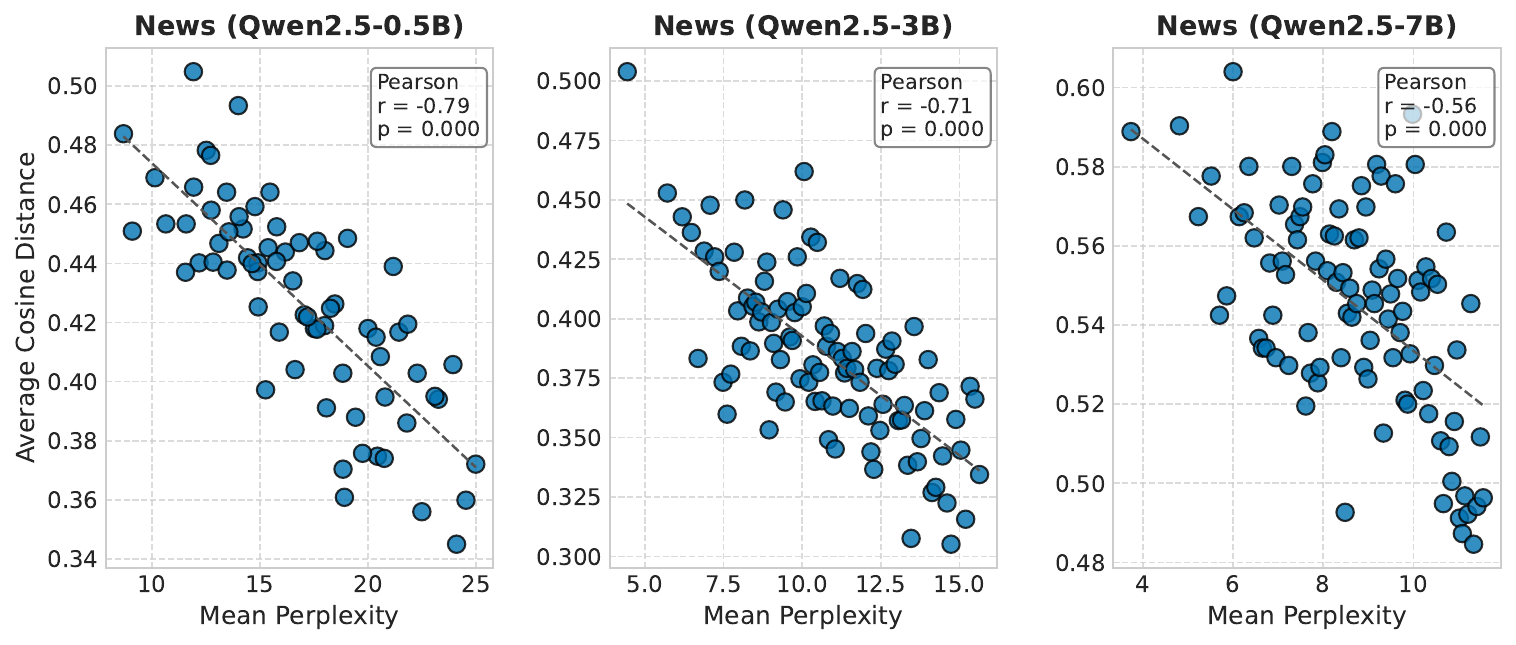}
    \caption{Perplexity vs.\ Average Pairwise Cosine Distance on CNN DailyMail (Qwen family).}
    \label{fig:appendix-cnn-qwen}
\end{figure}

\begin{figure}[h]
    \centering
    \includegraphics[width=0.45\textwidth]{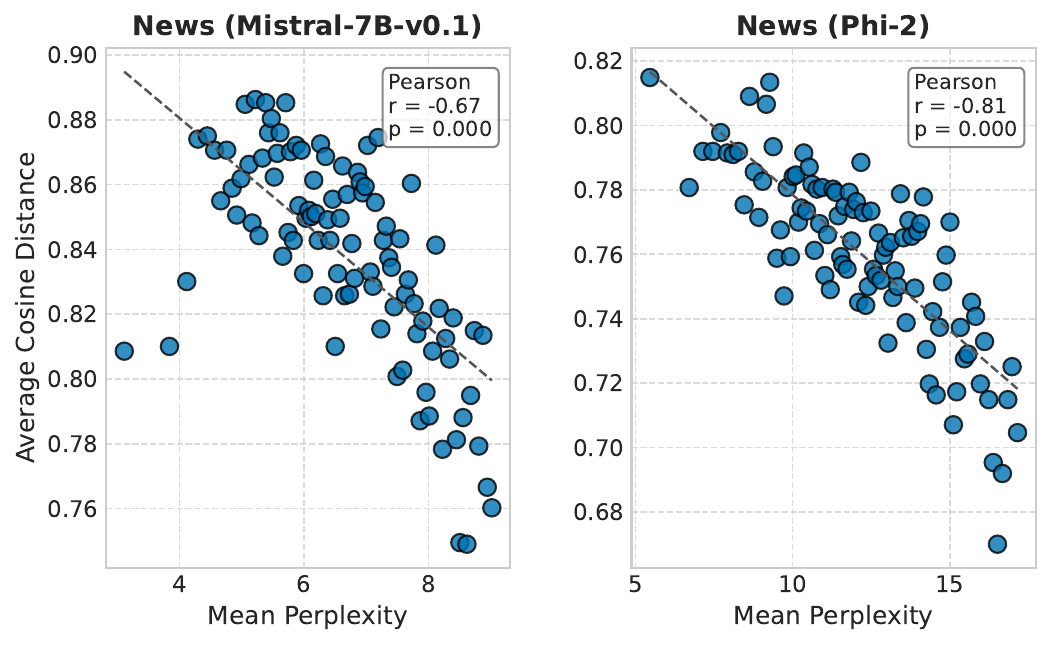}
    \caption{Perplexity vs.\ Average Pairwise Cosine Distance on CNN DailyMail (Mistral, Phi).}
    \label{fig:appendix-cnn-phi-mistral}
\end{figure}

\begin{figure}[h]
    \centering
    \includegraphics[width=0.65\textwidth]{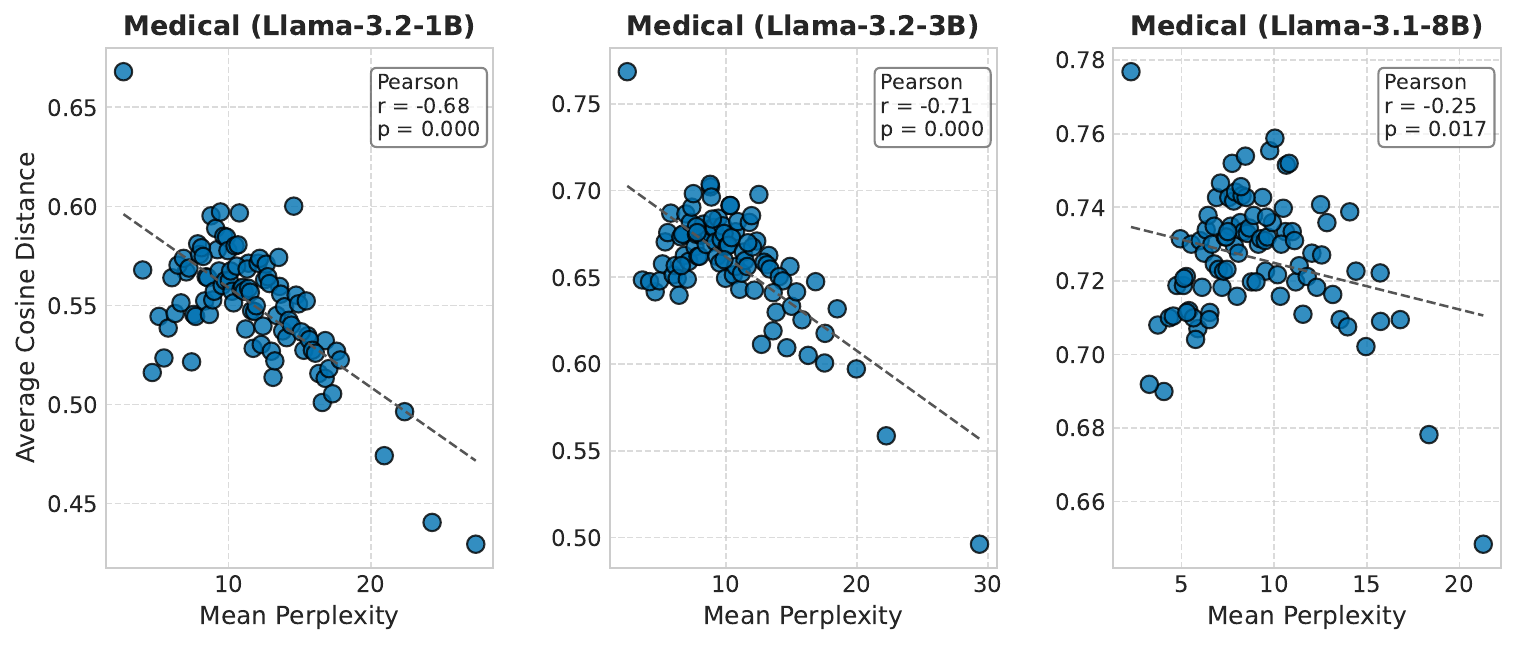}
    \caption{Perplexity vs.\ Average Pairwise Cosine Distance on PubMed Summarization (Llama family).}
    \label{fig:appendix-pubmed-llama}
\end{figure}

\begin{figure}[h]
    \centering
    \includegraphics[width=0.45\textwidth]{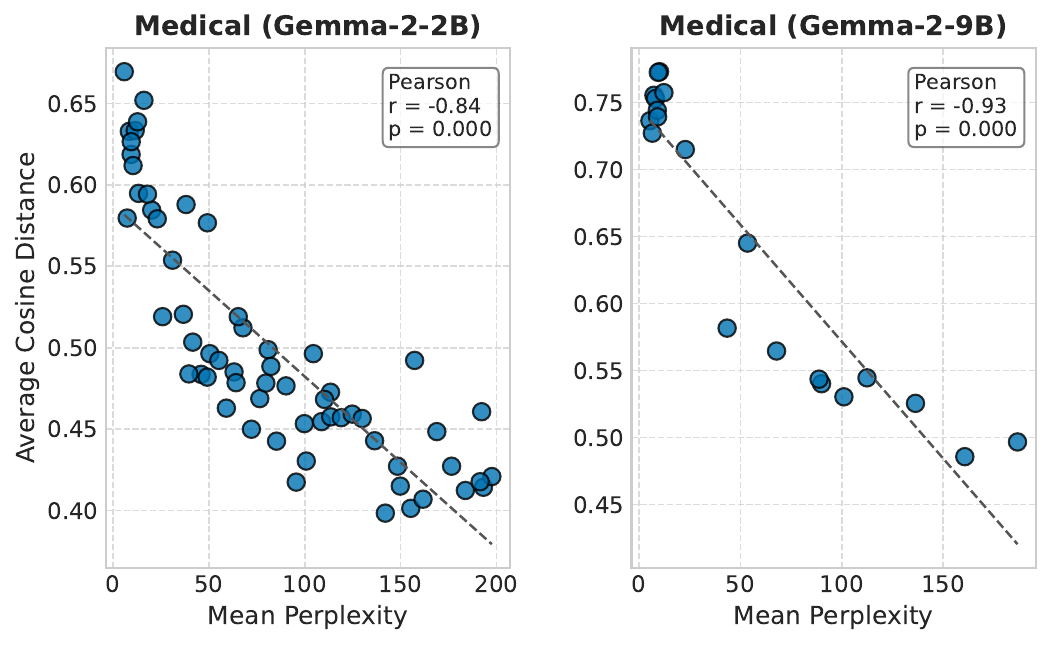}
    \caption{Perplexity vs.\ Average Pairwise Cosine Distance on PubMed Summarization (Gemma family).}
    \label{fig:appendix-pubmed-gemma}
\end{figure}

\begin{figure}[h]
    \centering
    \includegraphics[width=0.65\textwidth]{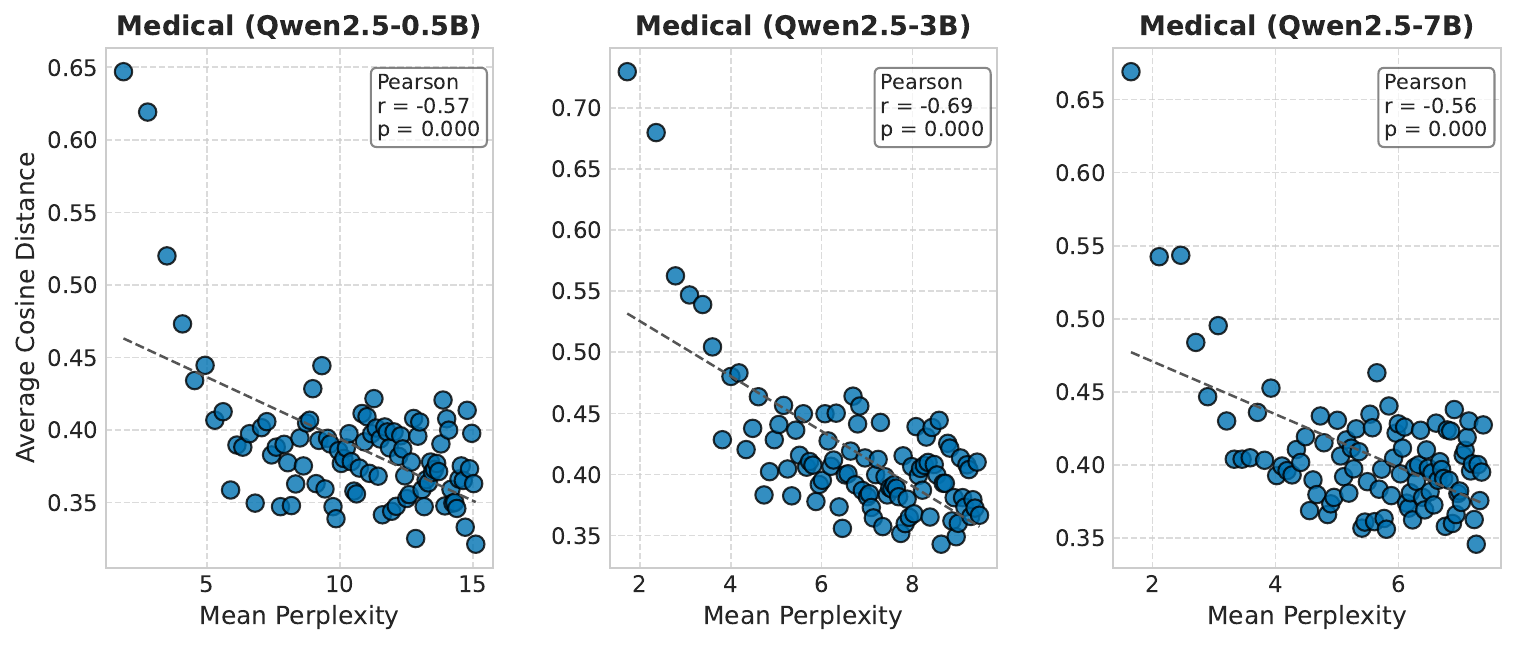}
    \caption{Perplexity vs.\ Average Pairwise Cosine Distance on PubMed Summarization (Qwen family).}
    \label{fig:appendix-pubmed-qwen}
\end{figure}

\begin{figure}[h]
    \centering
    \includegraphics[width=0.45\textwidth]{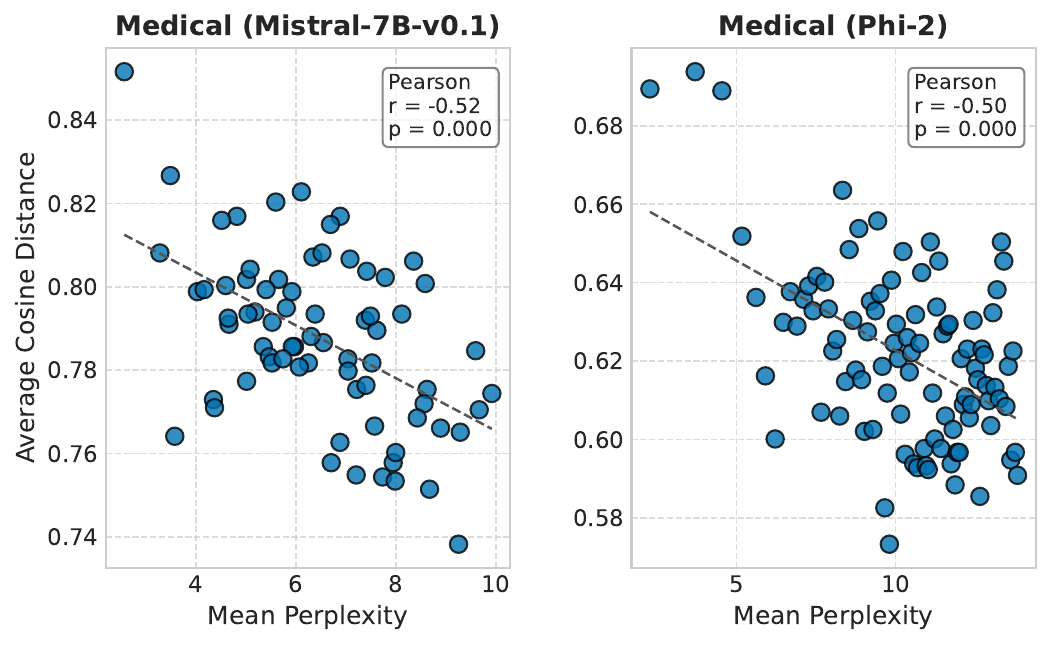}
    \caption{Perplexity vs.\ Average Pairwise Cosine Distance on PubMed Summarization (Mistral, Phi).}
    \label{fig:appendix-pubmed-phi-mistral}
\end{figure}

\clearpage

As shown in the figures, the observed \emph{negative correlation} between sequence-level perplexity and representation dispersion holds consistently across:
\begin{itemize}
    \item Multiple model sizes and architectures (Llama, Gemma, Mistral, Phi, Qwen).
    \item Multiple data domains (Wikitext-103, CNN DailyMail, PubMed Summarization).
\end{itemize}
These findings support the main paper's claim that \emph{lower-perplexity} contexts tend to occupy more ``spread out'' regions in the final-layer embedding space, while \emph{higher-perplexity} (i.e., more challenging) contexts appear more compressed.

\clearpage

\subsection{Details regarding Fine-Tuning Effect Experiments}
\label{sec:appendix-finetune}

In \S\ref{subsec:global-observations}, we examined how fine-tuning influenced representation dispersion. Below are the hyperparameters for the two fine-tuned LLaMA-3.2-1B models used in our experiments. We fine-tuned both checkpoints using the open-source LLaMA-Factory framework \footnote{\url{https://github.com/hiyouga/LLaMA-Factory}}.

\subsubsection{LoRA Fine-Tuned Model}
This model is a fine-tuned version of \href{https://huggingface.co/meta-llama/Llama-3.2-1B}{\texttt{meta-llama/Llama-3.2-1B}} on the Wikitext-103 dataset. 
It achieved the following on the evaluation set:
\begin{itemize}
    \item \textbf{Loss:} 2.1764
\end{itemize}

\paragraph{Training Hyperparameters.} 
\begin{itemize}
    \item \texttt{learning\_rate}: 0.0001
    \item \texttt{train\_batch\_size}: 8
    \item \texttt{eval\_batch\_size}: 1
    \item \texttt{seed}: 42
    \item \texttt{gradient\_accumulation\_steps}: 8
    \item \texttt{total\_train\_batch\_size}: 64
    \item \texttt{optimizer}: \texttt{adamw\_torch} with \(\beta_1=0.9\), \(\beta_2=0.999\), \(\epsilon = 1\times10^{-8}\), no additional arguments
    \item \texttt{lr\_scheduler\_type}: \texttt{cosine}
    \item \texttt{lr\_scheduler\_warmup\_ratio}: 0.1
    \item \texttt{num\_epochs}: 1.0
\end{itemize}

\subsubsection{Full-Parameter Fine-Tuned Model}
This model is also a fine-tuned version of \href{https://huggingface.co/meta-llama/Llama-3.2-1B}{\texttt{meta-llama/Llama-3.2-1B}} on the Wikitext-103 dataset. 
It achieved the following on the evaluation set:
\begin{itemize}
    \item \textbf{Loss:} 2.1333
\end{itemize}

\paragraph{Training Hyperparameters.}
\begin{itemize}
    \item \texttt{learning\_rate}: 1e-05
    \item \texttt{train\_batch\_size}: 2
    \item \texttt{eval\_batch\_size}: 1
    \item \texttt{seed}: 42
    \item \texttt{distributed\_type}: \texttt{multi-GPU}
    \item \texttt{num\_devices}: 2
    \item \texttt{gradient\_accumulation\_steps}: 16
    \item \texttt{total\_train\_batch\_size}: 64
    \item \texttt{total\_eval\_batch\_size}: 2
    \item \texttt{optimizer}: \texttt{Adam} with \(\beta_1=0.9\), \(\beta_2=0.999\), \(\epsilon=1\times10^{-8}\)
    \item \texttt{lr\_scheduler\_type}: \texttt{cosine}
    \item \texttt{lr\_scheduler\_warmup\_ratio}: 0.1
    \item \texttt{num\_epochs}: 5.0
\end{itemize}

\clearpage

\subsection{Details regarding Dispersion Within Semantic Clusters Training Hyperparameters}
\label{sec:appendix-semantic-cluster}

In \S\ref{subsec:semantic-dispersion}, we examined how dispersion evolves within carefully constructed \emph{semantic clusters} of text segments that share the same 10-gram continuation. Below are the training hyperparameters for the model used in this experiment:

\begin{itemize}
    \item \texttt{learning\_rate}: 1e-05
    \item \texttt{train\_batch\_size}: 10
    \item \texttt{eval\_batch\_size}: 1
    \item \texttt{seed}: 42
    \item \texttt{distributed\_type}: \texttt{multi-GPU}
    \item \texttt{num\_devices}: 8
    \item \texttt{gradient\_accumulation\_steps}: 8
    \item \texttt{total\_train\_batch\_size}: 640
    \item \texttt{total\_eval\_batch\_size}: 8
    \item \texttt{optimizer}: \texttt{ADAMW\_TORCH} with \(\beta_1=0.9\), \(\beta_2=0.999\), \(\epsilon=1\times10^{-8}\), no additional arguments
    \item \texttt{lr\_scheduler\_type}: \texttt{cosine}
    \item \texttt{lr\_scheduler\_warmup\_ratio}: 0.1
    \item \texttt{num\_epochs}: 5.0
\end{itemize}

\noindent We used these hyperparameters to train the model from a checkpoint of \texttt{meta-llama/Llama-3.2-1B} on WikiText-103, then tracked within-cluster and between-cluster distances of the resulting contextual embeddings at several checkpoints during training.  The model is also fine-tuned using the open-source LLaMA-Factory framework.

\clearpage

\subsection{Generalization to Held-Out Data}
\label{app:train-test-split}

To verify that the correlation between dispersion and perplexity holds on unseen data, we compare the standard distribution (Validation) against a strict Held-Out (Test) split using \textsc{Llama-3.2-1B}.

\begin{figure}[h]
    \centering
    \begin{subfigure}{0.48\textwidth}
        \centering
        \includegraphics[width=\linewidth]{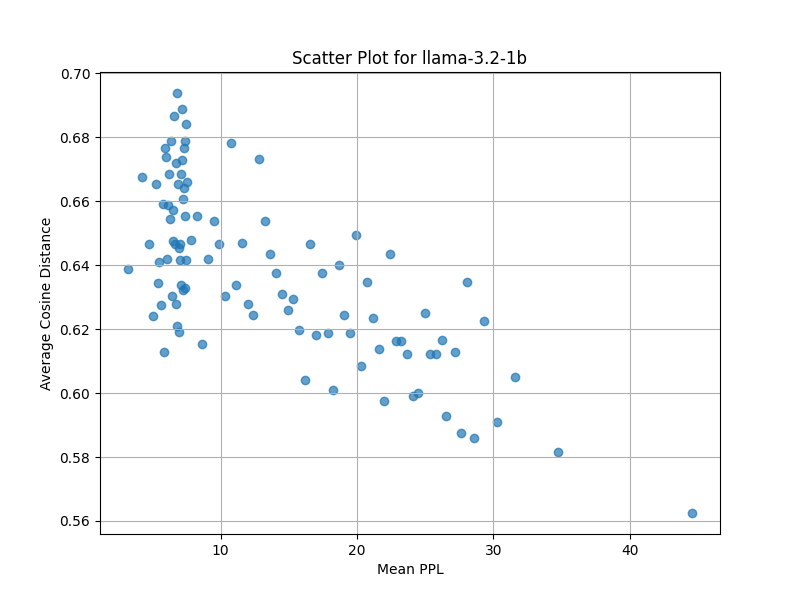}
        \caption{Validation Set (Standard)}
        \label{fig:gen-val}
    \end{subfigure}
    \hfill
    \begin{subfigure}{0.48\textwidth}
        \centering
        \includegraphics[width=\linewidth]{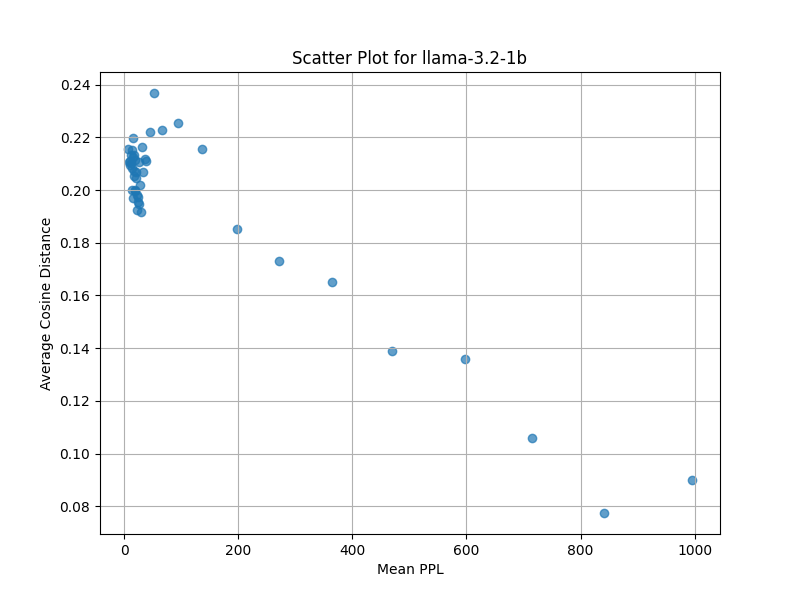}
        \caption{Validation vs. Test Split}
        \label{fig:gen-test}
    \end{subfigure}
    \caption{\textbf{Generalization Check (\textsc{Llama-3.2-1B}).} 
    Left: The standard correlation on the validation set. 
    Right: The correlation on the held-out test set (orange) vs validation (blue). The negative slope remains robust on unseen data.}
    \label{fig:train_test_split}
\end{figure}

As shown in \Cref{fig:train_test_split}, the negative correlation is consistent across both settings. While the test split (Right) shows higher perplexity overall, the geometric relationship remains: lower perplexity samples are associated with higher embedding dispersion.

\clearpage

\subsection{Vendi Score Analysis}
\label{app:vendi-score}

\begin{figure}[p] 
    \centering
    \begin{subfigure}{0.48\textwidth}
        \centering
        \includegraphics[width=\linewidth]{figures/cosine_llama-3.2-1b_wikitext.png}
        \caption{Cosine Distance (\textsc{Llama-3.2-1B})}
    \end{subfigure}
    \hfill
    \begin{subfigure}{0.48\textwidth}
        \centering
        \includegraphics[width=\linewidth]{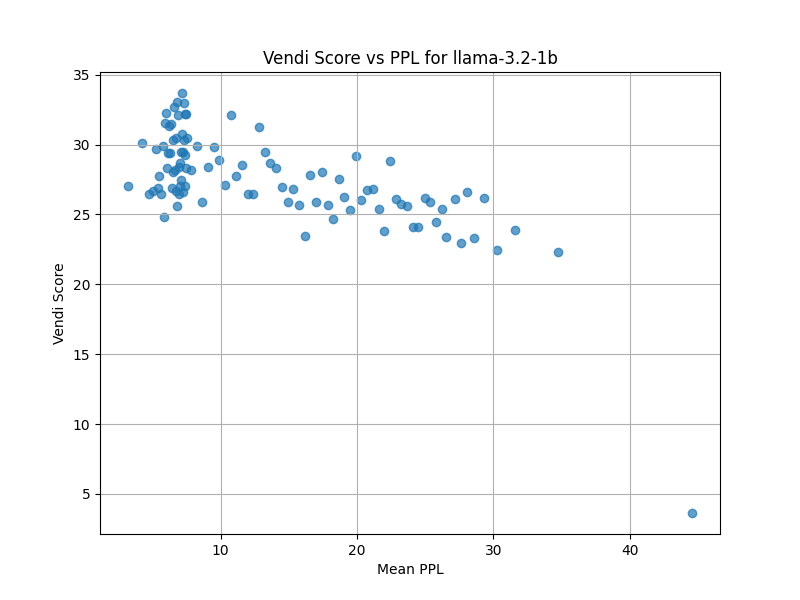}
        \caption{Vendi Score (\textsc{Llama-3.2-1B})}
    \end{subfigure}
    
    \vspace{1em}
    
    \begin{subfigure}{0.48\textwidth}
        \centering
        \includegraphics[width=\linewidth]{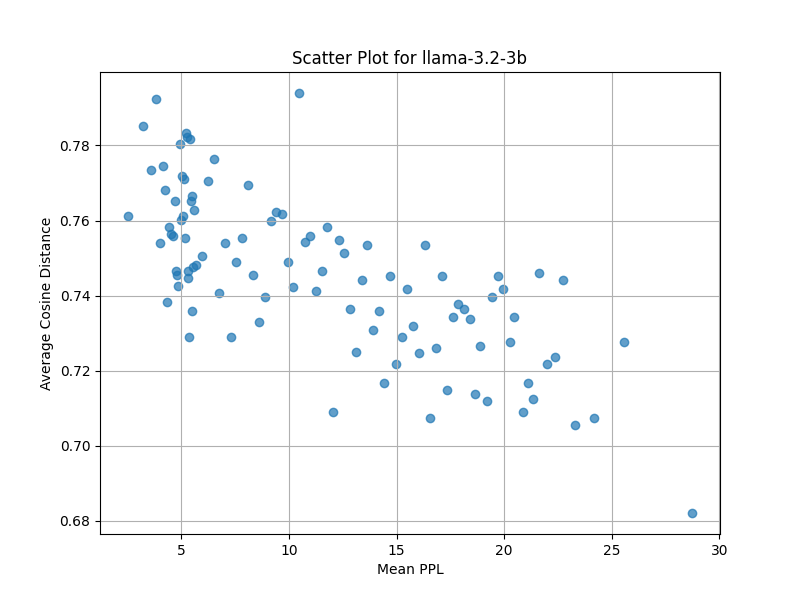}
        \caption{Cosine Distance (\textsc{Llama-3.2-3B})}
    \end{subfigure}
    \hfill
    \begin{subfigure}{0.48\textwidth}
        \centering
        \includegraphics[width=\linewidth]{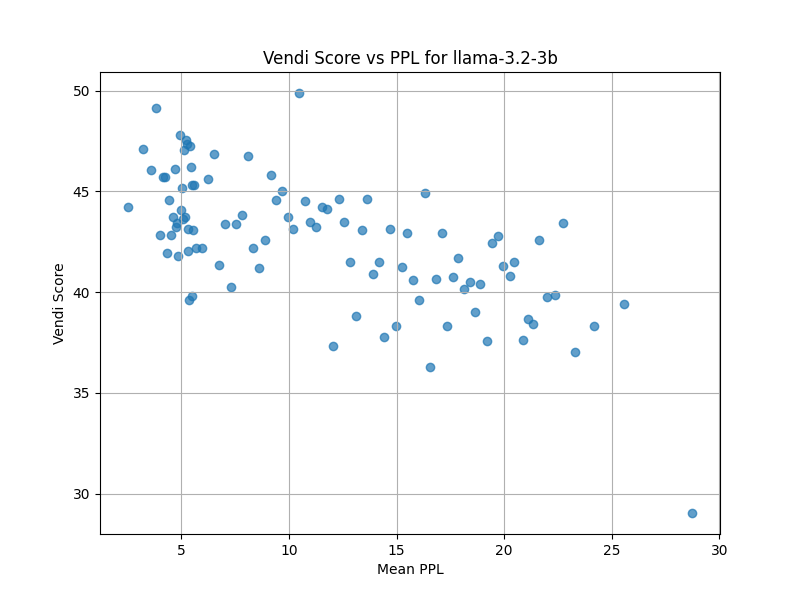}
        \caption{Vendi Score (\textsc{Llama-3.2-3B})}
    \end{subfigure}
    
    \vspace{1em}
    
    \begin{subfigure}{0.48\textwidth}
        \centering
        \includegraphics[width=\linewidth]{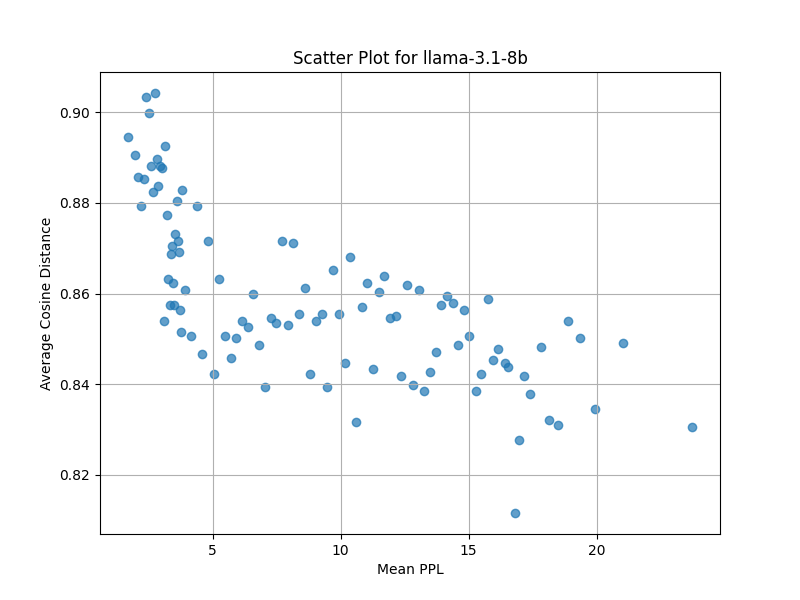}
        \caption{Cosine Distance (\textsc{Llama-3.1-8B})}
    \end{subfigure}
    \hfill
    \begin{subfigure}{0.48\textwidth}
        \centering
        \includegraphics[width=\linewidth]{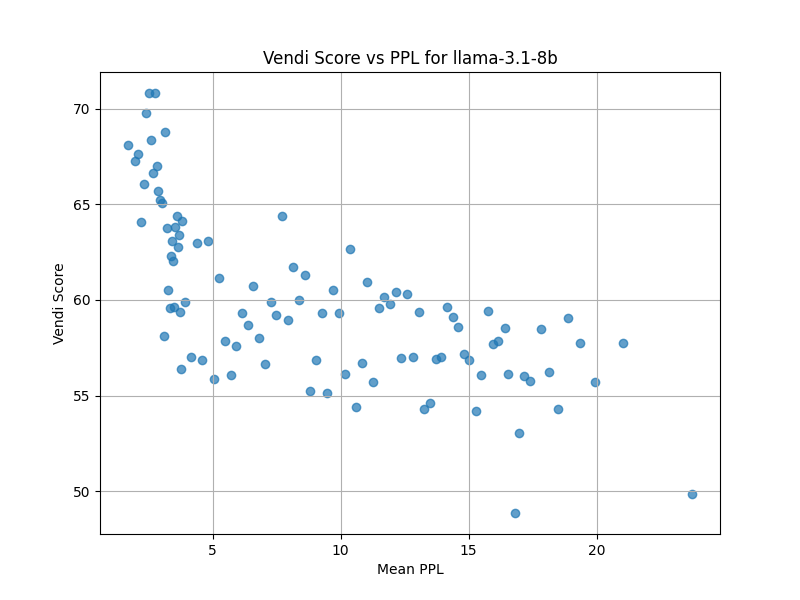}
        \caption{Vendi Score (\textsc{Llama-3.1-8B})}
    \end{subfigure}
    
    \caption{\textbf{Metric Robustness across Model Scales.} 
    We compare our primary metric (Average Cosine Distance, left) with the Vendi Score (right) across 1B, 3B, and 8B parameter models. The structural relationship with perplexity is identical across both metrics, validating the robustness of our geometric analysis.}
    \label{fig:vendi_robustness}
\end{figure}

To ensure our results are not specific to the Cosine Distance metric, we replicate our analysis using the \textbf{Vendi Score} \citep{friedman2023vendiscorediversityevaluation}. \Cref{fig:vendi_robustness} presents a side-by-side comparison of Average Cosine Distance (Left Column) and Vendi Score (Right Column) across three model scales.

In all cases (\textsc{Llama-3.2-1B}, \textsc{3.2-3B}, and \textsc{3.1-8B}), the Vendi Score follows the exact same trend as Cosine Distance: segments with lower perplexity exhibit higher diversity scores. This confirms that our main metric is a reliable proxy for the intrinsic dimensionality and spread of the representation space.

\color{black}

\clearpage

\section{Supplemental Materials for Applications of Representation Dispersion}
\label{app:application_supplemental}

\subsection{Details Regarding Ranking Example Hardness Without Labeled Data}
\label{app:downstream}

\subsubsection{Additional Results}
\label{app:acc_additional_res}

Below we provide extended experimental results following the methodology of \S\ref{sec:pred-downstream}. 
Each figure contains results for three models: \textbf{Llama-3.2-1B-Instruct}, \textbf{Llama-3.2-3B-Instruct}, and \textbf{Llama-3.1-8B-Instruct}. 

\begin{figure}[h!]
    \centering
    \includegraphics[width=1\linewidth]{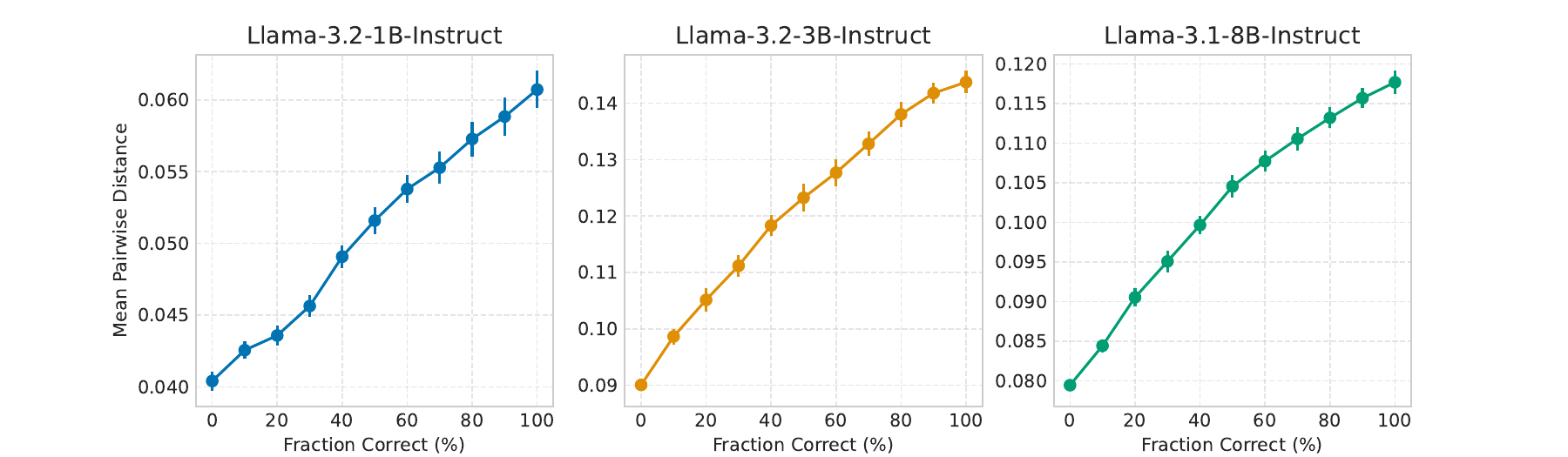}
    \caption{Downstream performance estimation on ARC Challenge 
    (containing results for Llama-3.2-1B-Instruct, Llama-3.2-3B-Instruct, and Llama-3.1-8B-Instruct).}
    \label{fig:arc-challenge}
\end{figure}

\begin{figure}[h!]
    \centering
    \includegraphics[width=1\linewidth]{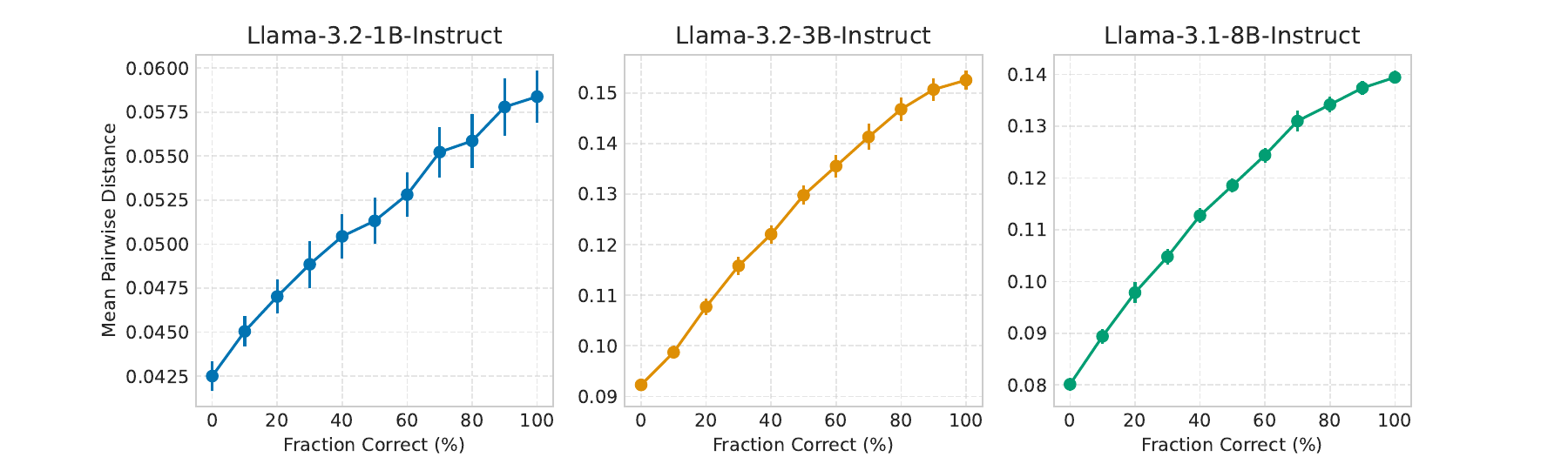}
    \caption{Downstream performance estimation on MMLU (English) 
    (containing results for Llama-3.2-1B-Instruct, Llama-3.2-3B-Instruct, and Llama-3.1-8B-Instruct).}
    \label{fig:mmlu}
\end{figure}

\begin{figure}[h!]
    \centering
    \includegraphics[width=1\linewidth]{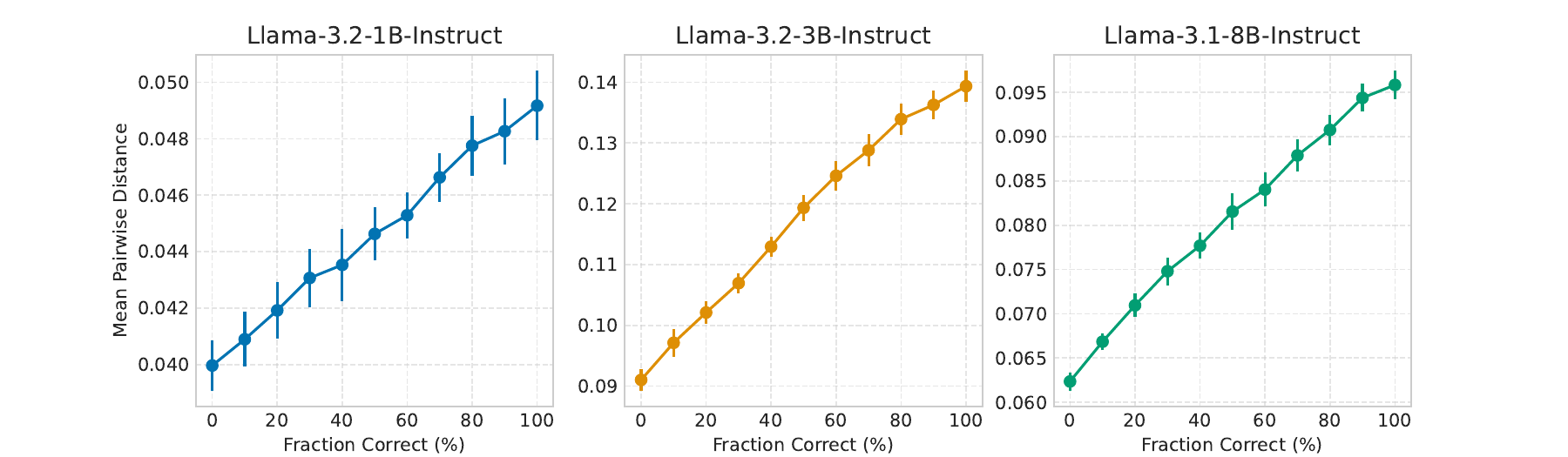}
    \caption{Downstream performance estimation on Multilingual MMLU (German) 
    (containing results for Llama-3.2-1B-Instruct, Llama-3.2-3B-Instruct, and Llama-3.1-8B-Instruct).}
    \label{fig:mmlu-de}
\end{figure}

\begin{figure}[h!]
    \centering
    \includegraphics[width=1\linewidth]{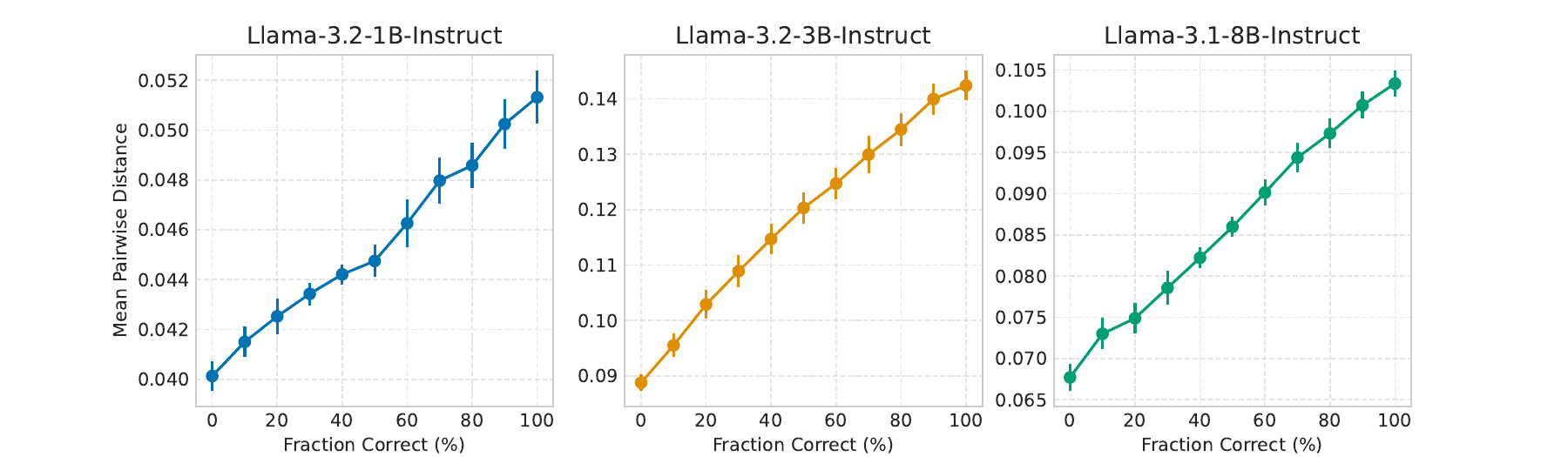}
    \caption{Downstream performance estimation on Multilingual MMLU (Spanish) 
    (containing results for Llama-3.2-1B-Instruct, Llama-3.2-3B-Instruct, and Llama-3.1-8B-Instruct).}
    \label{fig:mmlu-es}
\end{figure}

\begin{figure}[h!]
    \centering
    \includegraphics[width=1\linewidth]{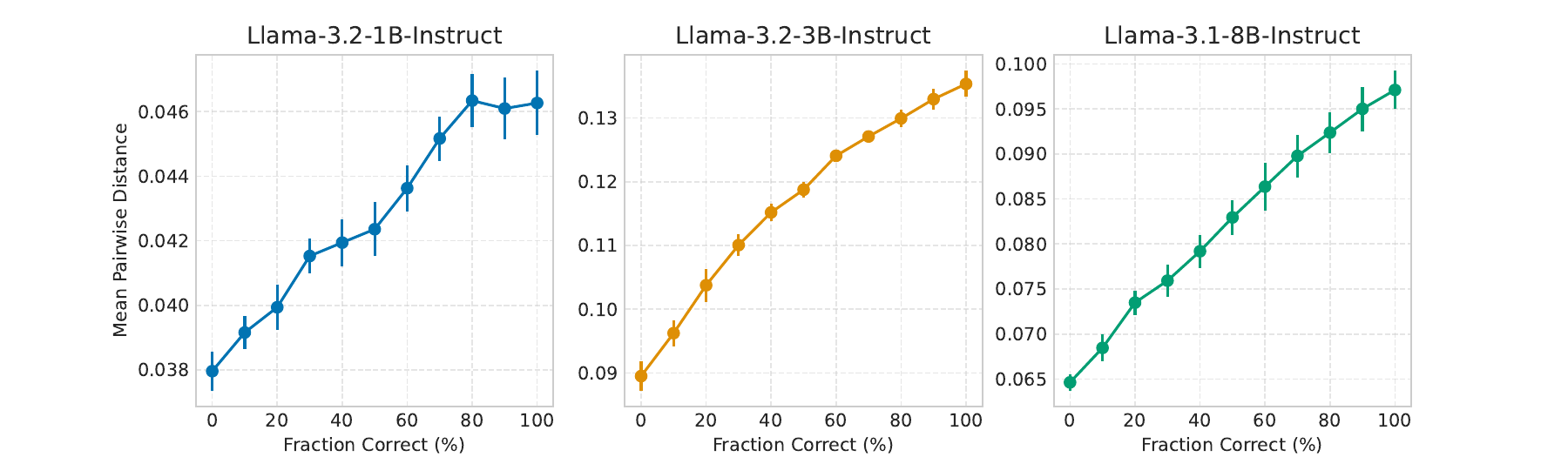}
    \caption{Downstream performance estimation on Multilingual MMLU (French) 
    (containing results for Llama-3.2-1B-Instruct, Llama-3.2-3B-Instruct, and Llama-3.1-8B-Instruct).}
    \label{fig:mmlu-fr}
\end{figure}

\begin{figure}[h!]
    \centering
    \includegraphics[width=1\linewidth]{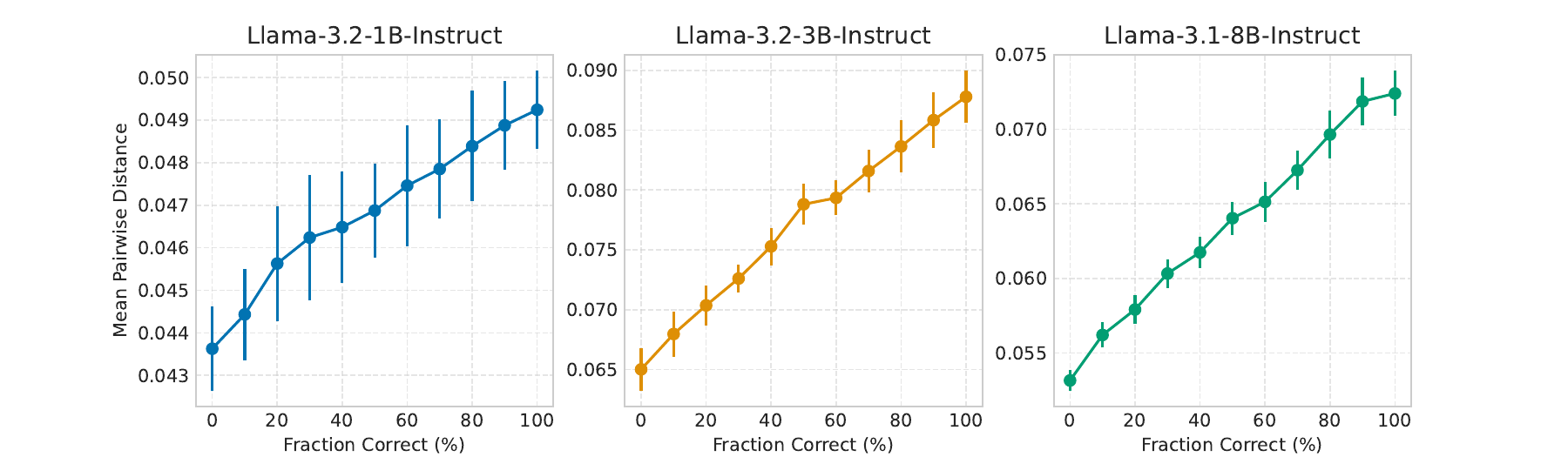}
    \caption{Downstream performance estimation on Multilingual MMLU (Hindi) 
    (containing results for Llama-3.2-1B-Instruct, Llama-3.2-3B-Instruct, and Llama-3.1-8B-Instruct).}
    \label{fig:mmlu-hi}
\end{figure}

\begin{figure}[h!]
    \centering
    \includegraphics[width=1\linewidth]{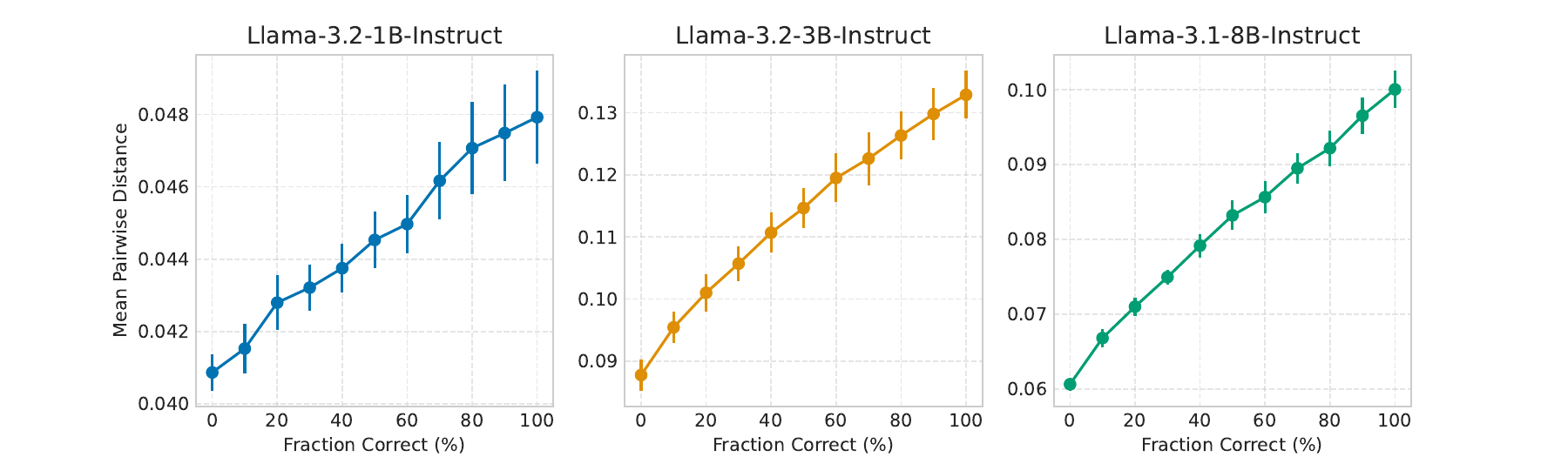}
    \caption{Downstream performance estimation on Multilingual MMLU (Italian) 
    (containing results for Llama-3.2-1B-Instruct, Llama-3.2-3B-Instruct, and Llama-3.1-8B-Instruct).}
    \label{fig:mmlu-it}
\end{figure}

\begin{figure}[h!]
    \centering
    \includegraphics[width=1\linewidth]{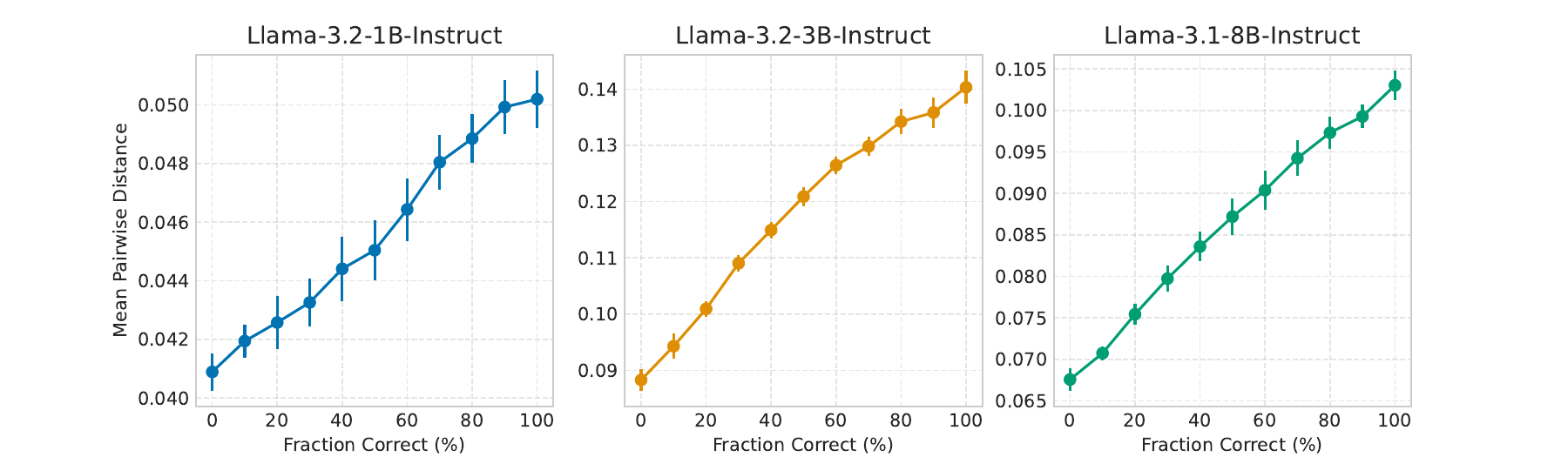}
    \caption{Downstream performance estimation on Multilingual MMLU (Portuguese) 
    (containing results for Llama-3.2-1B-Instruct, Llama-3.2-3B-Instruct, and Llama-3.1-8B-Instruct).}
    \label{fig:mmlu-pt}
\end{figure}

\begin{figure}[h!]
    \centering
    \includegraphics[width=1\linewidth]{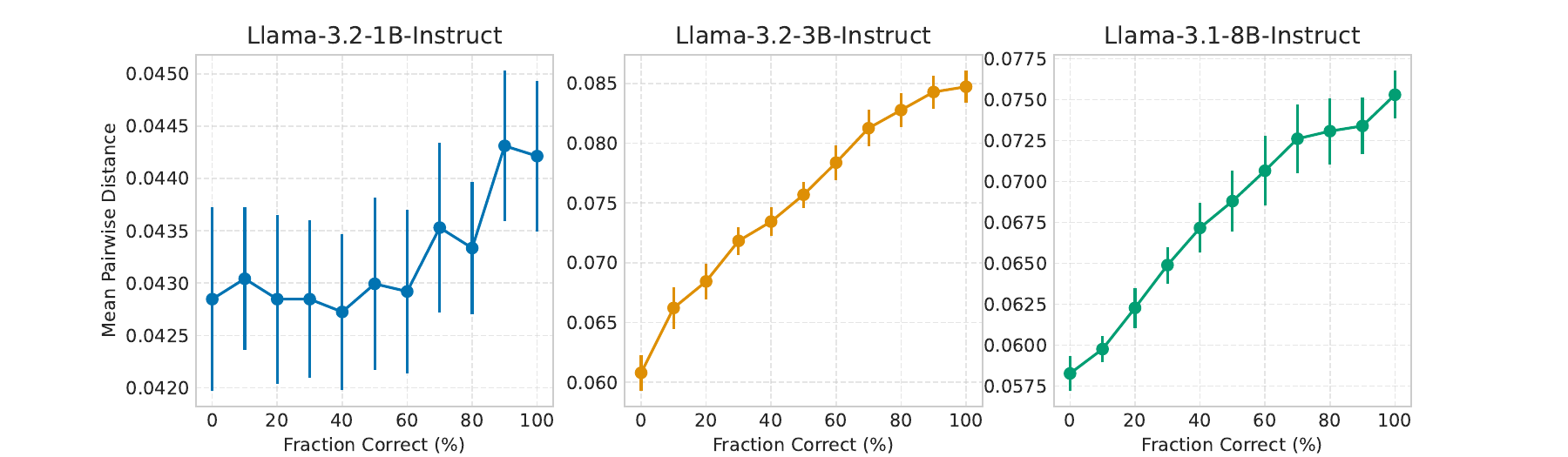}
    \caption{Downstream performance estimation on Multilingual MMLU (Thai) 
    (containing results for Llama-3.2-1B-Instruct, Llama-3.2-3B-Instruct, and Llama-3.1-8B-Instruct).}
    \label{fig:mmlu-th}
\end{figure}

\clearpage

\subsubsection{Comparison to Perplexity as a Hardness Signal}
\label{app:ppl-comparison}

In \S\ref{sec:pred-downstream} we used representation dispersion to rank slices of examples by their hardness for a fixed model and dataset. Here we replicate the same controlled slicing protocol but compare dispersion against the model's own perplexity on the query text as two competing label-free hardness signals. For each dataset (MMLU and \textsc{ARC-Challenge}) and each model size (Llama-3.2-1B/3B-Instruct and Llama-3.1-8B-Instruct), we construct synthetic slices with target fractions of correct answers from 0\% to 100\% in steps of 10\%. For every slice, we compute (i) the mean dispersion of the slice and (ii) the mean perplexity of the slice, averaging over 10 random seeds.

\begin{table}[h]
\centering
\small
\begin{tabular}{cccc}
\toprule
\textbf{Model} & \textbf{Frac.\ correct (\%)} & \textbf{Perplexity} & \textbf{Dispersion} \\
\midrule
1B & 100 & $5.273 \pm 0.063$ & $0.0584 \pm 0.0015$ \\
   & 90  & $5.232 \pm 0.056$ & $0.0578 \pm 0.0016$ \\
   & 80  & $5.196 \pm 0.068$ & $0.0559 \pm 0.0016$ \\
   & 70  & $5.201 \pm 0.064$ & $0.0553 \pm 0.0015$ \\
   & 60  & $5.178 \pm 0.050$ & $0.0528 \pm 0.0013$ \\
   & 50  & $5.166 \pm 0.033$ & $0.0513 \pm 0.0013$ \\
   & 40  & $5.147 \pm 0.044$ & $0.0504 \pm 0.0013$ \\
   & 30  & $5.092 \pm 0.054$ & $0.0488 \pm 0.0013$ \\
   & 20  & $5.038 \pm 0.042$ & $0.0470 \pm 0.0010$ \\
   & 10  & $5.070 \pm 0.032$ & $0.0450 \pm 0.0008$ \\
   & 0   & $5.058 \pm 0.033$ & $0.0425 \pm 0.0008$ \\
\midrule
3B & 100 & $4.740 \pm 0.055$ & $0.1525 \pm 0.0018$ \\
   & 90  & $4.753 \pm 0.043$ & $0.1507 \pm 0.0023$ \\
   & 80  & $4.761 \pm 0.044$ & $0.1468 \pm 0.0023$ \\
   & 70  & $4.738 \pm 0.040$ & $0.1414 \pm 0.0026$ \\
   & 60  & $4.730 \pm 0.041$ & $0.1355 \pm 0.0022$ \\
   & 50  & $4.760 \pm 0.050$ & $0.1298 \pm 0.0019$ \\
   & 40  & $4.803 \pm 0.049$ & $0.1221 \pm 0.0018$ \\
   & 30  & $4.780 \pm 0.042$ & $0.1158 \pm 0.0018$ \\
   & 20  & $4.742 \pm 0.033$ & $0.1077 \pm 0.0016$ \\
   & 10  & $4.762 \pm 0.040$ & $0.0987 \pm 0.0013$ \\
   & 0   & $4.751 \pm 0.025$ & $0.0923 \pm 0.0012$ \\
\midrule
8B & 100 & $4.528 \pm 0.044$ & $0.1395 \pm 0.0013$ \\
   & 90  & $4.500 \pm 0.043$ & $0.1374 \pm 0.0014$ \\
   & 80  & $4.477 \pm 0.047$ & $0.1342 \pm 0.0015$ \\
   & 70  & $4.472 \pm 0.047$ & $0.1310 \pm 0.0020$ \\
   & 60  & $4.488 \pm 0.048$ & $0.1244 \pm 0.0014$ \\
   & 50  & $4.498 \pm 0.051$ & $0.1186 \pm 0.0014$ \\
   & 40  & $4.488 \pm 0.045$ & $0.1127 \pm 0.0014$ \\
   & 30  & $4.458 \pm 0.049$ & $0.1048 \pm 0.0016$ \\
   & 20  & $4.438 \pm 0.045$ & $0.0979 \pm 0.0020$ \\
   & 10  & $4.419 \pm 0.044$ & $0.0894 \pm 0.0015$ \\
   & 0   & $4.418 \pm 0.043$ & $0.0801 \pm 0.0012$ \\
\bottomrule
\end{tabular}
\caption{{\textsc{MMLU} controlled slices: mean perplexity (ppl) and dispersion as a function of the fraction of correct answers. Dispersion is strongly monotone with correctness for all model sizes, whereas perplexity is nearly flat.}}
\label{tab:mmlu_ppl_dispersion}
\end{table}

\begin{table}[h]
\centering
\small
\begin{tabular}{cccc}
\toprule
\textbf{Model} & \textbf{Frac.\ correct (\%)} & \textbf{Perplexity} & \textbf{Dispersion} \\
\midrule
1B & 100 & $21.846 \pm 0.083$ & $0.0607 \pm 0.0013$ \\
   & 90  & $21.670 \pm 0.105$ & $0.0589 \pm 0.0013$ \\
   & 80  & $21.466 \pm 0.148$ & $0.0573 \pm 0.0012$ \\
   & 70  & $21.289 \pm 0.130$ & $0.0553 \pm 0.0011$ \\
   & 60  & $21.389 \pm 0.183$ & $0.0538 \pm 0.0010$ \\
   & 50  & $21.094 \pm 0.226$ & $0.0516 \pm 0.0009$ \\
   & 40  & $20.898 \pm 0.236$ & $0.0491 \pm 0.0008$ \\
   & 30  & $20.842 \pm 0.225$ & $0.0456 \pm 0.0007$ \\
   & 20  & $20.779 \pm 0.239$ & $0.0436 \pm 0.0007$ \\
   & 10  & $20.561 \pm 0.237$ & $0.0426 \pm 0.0006$ \\
   & 0   & $20.589 \pm 0.227$ & $0.0404 \pm 0.0007$ \\
\midrule
3B & 100 & $13.943 \pm 0.090$ & $0.1437 \pm 0.0020$ \\
   & 90  & $13.917 \pm 0.099$ & $0.1418 \pm 0.0019$ \\
   & 80  & $13.875 \pm 0.101$ & $0.1381 \pm 0.0022$ \\
   & 70  & $13.957 \pm 0.111$ & $0.1328 \pm 0.0021$ \\
   & 60  & $13.933 \pm 0.107$ & $0.1277 \pm 0.0024$ \\
   & 50  & $14.052 \pm 0.084$ & $0.1232 \pm 0.0025$ \\
   & 40  & $13.966 \pm 0.091$ & $0.1183 \pm 0.0018$ \\
   & 30  & $13.939 \pm 0.120$ & $0.1111 \pm 0.0019$ \\
   & 20  & $13.915 \pm 0.117$ & $0.1051 \pm 0.0020$ \\
   & 10  & $13.862 \pm 0.081$ & $0.0986 \pm 0.0014$ \\
   & 0   & $13.846 \pm 0.090$ & $0.0900 \pm 0.0010$ \\
\midrule
8B & 100 & $16.313 \pm 0.157$ & $0.1177 \pm 0.0014$ \\
   & 90  & $16.272 \pm 0.160$ & $0.1157 \pm 0.0013$ \\
   & 80  & $16.247 \pm 0.156$ & $0.1132 \pm 0.0013$ \\
   & 70  & $16.102 \pm 0.189$ & $0.1105 \pm 0.0015$ \\
   & 60  & $15.939 \pm 0.176$ & $0.1077 \pm 0.0013$ \\
   & 50  & $16.052 \pm 0.158$ & $0.1045 \pm 0.0014$ \\
   & 40  & $15.931 \pm 0.166$ & $0.0997 \pm 0.0011$ \\
   & 30  & $15.869 \pm 0.109$ & $0.0951 \pm 0.0014$ \\
   & 20  & $15.748 \pm 0.113$ & $0.0905 \pm 0.0012$ \\
   & 10  & $15.755 \pm 0.147$ & $0.0844 \pm 0.0008$ \\
   & 0   & $15.835 \pm 0.094$ & $0.0794 \pm 0.0007$ \\
\bottomrule
\end{tabular}
\caption{\textsc{ARC-Challenge} controlled slices: mean perplexity (ppl) and dispersion as a function of the fraction of correct answers. Again, dispersion is strongly monotone with correctness, whereas perplexity fails to separate easy from hard slices.}
\label{tab:arc_ppl_dispersion}
\end{table}

Across both datasets and all three model sizes, dispersion is strongly monotone with the fraction of correct answers, while perplexity is nearly constant and sometimes even slightly better (lower) on the hardest slices than on the easiest ones. This supports the claim that dispersion captures geometric information about contextual separation that is not reducible to next-token confidence alone.

\clearpage

\subsubsection{Calibration by Dispersion}
\label{app:dispersion-calibration}

To test whether dispersion can be used as a calibrated proxy for accuracy on truly unseen data, we perform a ``Calibration by Dispersion'' experiment on MMLU. We partition the dataset into three disjoint subsets: (1) a \emph{Calibration Set} (20\% of the data), (2) a \emph{Validation Set} (10\%), and (3) a \emph{Test Set} (70\%). On the calibration set we use a dense sampling strategy to create synthetic batches with target accuracies ranging from 0\% to 100\% in 2\% increments, which allows us to densely sample the shape of the Dispersion$\to$Accuracy curve.

We then train a suite of regression models—linear, polynomial, isotonic, and random forest—to predict accuracy from dispersion, and select the model that minimizes prediction error on the validation set. Finally, we apply the selected regressor to the \emph{unseen} test set (using dispersion only) and compare the predicted accuracy to the true accuracy. We repeat this procedure over 10 random seeds and report the mean absolute error (MAE) between predicted and true accuracy. Table~\ref{tab:dispersion_calibration_mae} summarizes the results.

\begin{table}[h]
\centering
\small
\begin{tabular}{lc}
\toprule
\textbf{Model} & \textbf{MAE (\%)} \\
\midrule
Llama-3.2-1B-Instruct & 1.39 \\
Llama-3.2-3B-Instruct & 1.75 \\
Llama-3.1-8B-Instruct & 2.15 \\
\bottomrule
\end{tabular}
\caption{Mean Absolute Error (MAE) between predicted accuracy (derived solely from dispersion) and true accuracy on a held-out 70\% MMLU test set. Results are averaged over 10 random seeds. Dispersion supports highly accurate calibration, with absolute error of roughly 1.4--2.2 percentage points.}
\label{tab:dispersion_calibration_mae}
\end{table}

These results show that, for a fixed model and dataset, dispersion is not only monotonically related to correctness but can also be mapped to a well-calibrated accuracy estimate with small error. A practical workflow emerging from this experiment is:
\begin{enumerate}
    \item Label a small random subsample of the data (e.g., 10--20\%).
    \item Fit a standard regression curve (e.g., isotonic or low-degree polynomial) from Dispersion $\to$ Accuracy on this subsample.
    \item Apply the learned mapping to the remaining unlabeled data to estimate performance with a typical margin of error of about 2 percentage points.
\end{enumerate}
While this calibration is, by construction, model- and domain-specific, it demonstrates that dispersion can serve as a practical, calibrated proxy for accuracy once a small labeled calibration set is available.

\clearpage

\subsubsection{Testing Distribution Shift}
\label{app:token_disp_new}

To probe robustness under distribution shift, we further apply the controlled slicing experiment from \S\ref{sec:pred-downstream} to three datasets with distributions very different from MMLU and \textsc{ARC-Challenge}: \textsc{hellaswag\_chat}, \textsc{if\_eval}, and \textsc{quail}. For each dataset and for each of the three Llama-Instruct models, we construct slices with target fractions correct from 0\% to 100\% in steps of 10\%, and compute the mean dispersion (with standard errors) for each slice. In all cases we observe the same qualitative trend: higher fractions correct correspond to larger mean dispersion, confirming that dispersion remains a reliable hardness signal even under substantial distribution shift. The full numbers are reported in Tables~\ref{tab:hellaswag_dispersion}--\ref{tab:quail_dispersion}.

\begin{table}[t]
\centering
\small
\begin{tabular}{cccc}
\toprule
\textbf{Frac.\ correct (\%)} & \textbf{1B mean $\pm$ s.e.} & \textbf{3B mean $\pm$ s.e.} & \textbf{8B mean $\pm$ s.e.} \\
\midrule
100 & $0.0350 \pm 0.0004$ & $0.0709 \pm 0.0010$ & $0.0750 \pm 0.0012$ \\
90  & $0.0345 \pm 0.0005$ & $0.0694 \pm 0.0009$ & $0.0738 \pm 0.0013$ \\
80  & $0.0343 \pm 0.0004$ & $0.0674 \pm 0.0010$ & $0.0714 \pm 0.0012$ \\
70  & $0.0341 \pm 0.0005$ & $0.0665 \pm 0.0009$ & $0.0692 \pm 0.0011$ \\
60  & $0.0335 \pm 0.0006$ & $0.0642 \pm 0.0011$ & $0.0669 \pm 0.0009$ \\
50  & $0.0325 \pm 0.0006$ & $0.0620 \pm 0.0011$ & $0.0646 \pm 0.0007$ \\
40  & $0.0319 \pm 0.0006$ & $0.0599 \pm 0.0009$ & $0.0627 \pm 0.0006$ \\
30  & $0.0311 \pm 0.0006$ & $0.0583 \pm 0.0010$ & $0.0605 \pm 0.0006$ \\
20  & $0.0306 \pm 0.0006$ & $0.0560 \pm 0.0007$ & $0.0578 \pm 0.0009$ \\
10  & $0.0306 \pm 0.0006$ & $0.0540 \pm 0.0006$ & $0.0546 \pm 0.0010$ \\
0   & $0.0302 \pm 0.0005$ & $0.0511 \pm 0.0008$ & $0.0513 \pm 0.0008$ \\
\bottomrule
\end{tabular}
\caption{\textsc{hellaswag\_chat}: mean dispersion (with standard error) as a function of the fraction of correct answers for Llama-3.2-1B/3B-Instruct and Llama-3.1-8B-Instruct.}
\label{tab:hellaswag_dispersion}
\end{table}

\begin{table}[t]
\centering
\small
\begin{tabular}{cccc}
\toprule
\textbf{Frac.\ correct (\%)} & \textbf{1B mean $\pm$ s.e.} & \textbf{3B mean $\pm$ s.e.} & \textbf{8B mean $\pm$ s.e.} \\
\midrule
100 & $0.7289 \pm 0.0069$ & $0.7266 \pm 0.0064$ & $0.7113 \pm 0.0039$ \\
90  & $0.7277 \pm 0.0068$ & $0.7277 \pm 0.0070$ & $0.7141 \pm 0.0043$ \\
80  & $0.7238 \pm 0.0061$ & $0.7250 \pm 0.0048$ & $0.7094 \pm 0.0031$ \\
70  & $0.7203 \pm 0.0054$ & $0.7270 \pm 0.0050$ & $0.7074 \pm 0.0031$ \\
60  & $0.7180 \pm 0.0052$ & $0.7246 \pm 0.0035$ & $0.7039 \pm 0.0037$ \\
50  & $0.7145 \pm 0.0053$ & $0.7211 \pm 0.0038$ & $0.7043 \pm 0.0055$ \\
40  & $0.7133 \pm 0.0042$ & $0.7238 \pm 0.0033$ & $0.7031 \pm 0.0048$ \\
30  & $0.7113 \pm 0.0045$ & $0.7191 \pm 0.0042$ & $0.7031 \pm 0.0036$ \\
20  & $0.7039 \pm 0.0037$ & $0.7156 \pm 0.0056$ & $0.6996 \pm 0.0041$ \\
10  & $0.7070 \pm 0.0030$ & $0.7172 \pm 0.0044$ & $0.6961 \pm 0.0021$ \\
0   & $0.7078 \pm 0.0018$ & $0.7164 \pm 0.0039$ & $0.6926 \pm 0.0015$ \\
\bottomrule
\end{tabular}
\caption{\textsc{if\_eval}: mean dispersion (with standard error) as a function of the fraction of correct answers for Llama-3.2-1B/3B-Instruct and Llama-3.1-8B-Instruct.}
\label{tab:if_eval_dispersion}
\end{table}

\begin{table}[t]
\centering
\small
\begin{tabular}{cccc}
\toprule
\textbf{Frac.\ correct (\%)} & \textbf{1B mean $\pm$ s.e.} & \textbf{3B mean $\pm$ s.e.} & \textbf{8B mean $\pm$ s.e.} \\
\midrule
100 & $0.1938 \pm 0.0020$ & $0.1259 \pm 0.0024$ & $0.0893 \pm 0.0014$ \\
90  & $0.1936 \pm 0.0020$ & $0.1234 \pm 0.0024$ & $0.0880 \pm 0.0015$ \\
80  & $0.1941 \pm 0.0021$ & $0.1220 \pm 0.0023$ & $0.0865 \pm 0.0014$ \\
70  & $0.1930 \pm 0.0021$ & $0.1207 \pm 0.0023$ & $0.0844 \pm 0.0014$ \\
60  & $0.1931 \pm 0.0023$ & $0.1186 \pm 0.0028$ & $0.0826 \pm 0.0014$ \\
50  & $0.1915 \pm 0.0021$ & $0.1165 \pm 0.0024$ & $0.0806 \pm 0.0012$ \\
40  & $0.1913 \pm 0.0016$ & $0.1148 \pm 0.0018$ & $0.0788 \pm 0.0010$ \\
30  & $0.1901 \pm 0.0013$ & $0.1122 \pm 0.0024$ & $0.0762 \pm 0.0007$ \\
20  & $0.1898 \pm 0.0012$ & $0.1087 \pm 0.0022$ & $0.0735 \pm 0.0009$ \\
10  & $0.1876 \pm 0.0011$ & $0.1046 \pm 0.0019$ & $0.0686 \pm 0.0009$ \\
0   & $0.1884 \pm 0.0012$ & $0.1014 \pm 0.0020$ & $0.0656 \pm 0.0009$ \\
\bottomrule
\end{tabular}
\caption{\textsc{quail}: mean dispersion (with standard error) as a function of the fraction of correct answers for Llama-3.2-1B/3B-Instruct and Llama-3.1-8B-Instruct.}
\label{tab:quail_dispersion}
\end{table}

\color{black}

\clearpage

\subsection{Details Regarding Representation Dispersion for Model Selection}
\label{app:dispersion-performance}

\subsubsection{Full Numeric Statistics}

This appendix compiles the full numeric statistics that underpin the analyses in
\S\ref{sec:embedding-dispersion-modelsel}.  We report complete Euclidean and
cosine distance figures for every model variant, together with their
task-specific accuracies, so that readers can perform fine-grained checks,
reproduce our correlation calculations, and explore alternative dispersion
metrics. \Cref{tab:qwen-euclidean-dispersion}, \Cref{tab:qwen-cosine-dispersion}, \Cref{tab:llama-code-euclidean-dispersion} and \Cref{tab:llama-code-cosine-dispersion} complement the visual summaries in
Figure~\ref{fig:math_code_dispersion} by exposing each component of the
\emph{Dispersion Gap} in detail.

\begin{table}[h!]
    \centering
    \footnotesize
    \caption{\textbf{Embedding Dispersion vs.\ MATH Performance (Qwen Variants, Euclidean).}
    We show the mean $\pm$ standard error of \textbf{Euclidean} distances among digit embeddings (D--D), among non-math tokens (NM--NM), and between digits and non-math tokens (D--NM). We also list each model’s accuracy on MATH (\%).}
    \label{tab:qwen-euclidean-dispersion}
    \resizebox{0.85\linewidth}{!}{
    \begin{tabular}{lcccc}
        \toprule
        \textbf{Model} 
        & \multicolumn{3}{c}{\textbf{Euclidean Distances}} 
        & \textbf{MATH (\%)} \\
        \cmidrule(lr){2-4}
         & D--D & NM--NM & D--NM & \\
        \midrule
        \textbf{Qwen2.5-1.5B}         & $0.7006 \pm 0.0087$ & $1.4072 \pm 0.0268$ & $1.4331 \pm 0.0222$ & 35.0 \\
        \textbf{Qwen2.5-Math-1.5B}    & $0.8916 \pm 0.0111$ & $1.6423 \pm 0.0286$ & $1.6991 \pm 0.0203$ & 49.8 \\
        \textbf{Distill-Qwen-1.5B}    & $0.9406 \pm 0.0103$ & $1.6104 \pm 0.0247$ & $1.7014 \pm 0.0188$ & 83.9 \\
        \midrule
        \textbf{Qwen2.5-7B}           & $0.4505 \pm 0.0052$ & $0.8712 \pm 0.0189$ & $0.9840 \pm 0.0115$ & 49.8 \\
        \textbf{Qwen2.5-Math-7B}      & $0.6896 \pm 0.0076$ & $1.3479 \pm 0.0287$ & $1.4047 \pm 0.0205$ & 55.4 \\
        \textbf{Distill-Qwen-7B}      & $0.7216 \pm 0.0076$ & $1.3406 \pm 0.0292$ & $1.4244 \pm 0.0215$ & 92.8 \\
        \midrule
        \textbf{Qwen2.5-14B}          & $0.6993 \pm 0.0074$ & $1.5284 \pm 0.0314$ & $1.4550 \pm 0.0168$ & 55.6 \\
        \textbf{Distill-Qwen-14B}     & $0.7415 \pm 0.0073$ & $1.5223 \pm 0.0402$ & $1.4659 \pm 0.0229$ & 93.9 \\
        \bottomrule
    \end{tabular}
    }
\end{table}

\begin{table}[h!]
    \centering
    \footnotesize
    \caption{\textbf{Embedding Dispersion vs.\ MATH Performance (Qwen Variants, Cosine).}
    We show the mean $\pm$ standard error of \textbf{Cosine} distances among digit embeddings (D--D), among non-math tokens (NM--NM), and between digits and non-math tokens (D--NM). We also list each model’s accuracy on MATH (\%).}
    \label{tab:qwen-cosine-dispersion}
    \resizebox{0.85\linewidth}{!}{
    \begin{tabular}{lcccc}
        \toprule
        \textbf{Model} 
        & \multicolumn{3}{c}{\textbf{Cosine Distances}} 
        & \textbf{MATH (\%)} \\
        \cmidrule(lr){2-4}
         & D--D & NM--NM & D--NM & \\
        \midrule
        \textbf{Qwen2.5-1.5B}         & $0.2489 \pm 0.0060$ & $0.9334 \pm 0.0089$ & $1.0068 \pm 0.0217$ & 35.0 \\
        \textbf{Qwen2.5-Math-1.5B}    & $0.3347 \pm 0.0074$ & $0.8984 \pm 0.0110$ & $1.0755 \pm 0.0172$ & 49.8 \\
        \textbf{Distill-Qwen-1.5B}    & $0.3562 \pm 0.0073$ & $0.8973 \pm 0.0109$ & $1.0781 \pm 0.0165$ & 83.9 \\
        \midrule
        \textbf{Qwen2.5-7B}           & $0.1786 \pm 0.0042$ & $0.9360 \pm 0.0072$ & $1.0000 \pm 0.0159$ & 49.8 \\
        \textbf{Qwen2.5-Math-7B}      & $0.2672 \pm 0.0061$ & $0.9257 \pm 0.0089$ & $1.0554 \pm 0.0166$ & 55.4 \\
        \textbf{Distill-Qwen-7B}      & $0.2775 \pm 0.0061$ & $0.9260 \pm 0.0084$ & $1.0640 \pm 0.0177$ & 92.8 \\
        \midrule
        \textbf{Qwen2.5-14B}          & $0.2573 \pm 0.0053$ & $0.9333 \pm 0.0084$ & $0.9622 \pm 0.0100$ & 55.6 \\
        \textbf{Distill-Qwen-14B}     & $0.2811 \pm 0.0054$ & $0.9350 \pm 0.0111$ & $0.9701 \pm 0.0135$ & 93.9 \\
        \bottomrule
    \end{tabular}
    }
\end{table}

\begin{table}[h!]
    \centering
    \footnotesize
    \caption{\textbf{Embedding Dispersion vs.\ HumanEval Performance (Llama2 vs.\ CodeLlama, Euclidean).}
    We report the mean $\pm$ standard error of \textbf{Euclidean} distances among code tokens (C--C), among non-code tokens (NC--NC), and between code and non-code tokens (C--NC). We also list each model’s HumanEval pass@1 (\%).}
    \label{tab:llama-code-euclidean-dispersion}
    \resizebox{0.88\linewidth}{!}{
    \begin{tabular}{lcccc}
        \toprule
        \textbf{Model} &
        \multicolumn{3}{c}{\textbf{Euclidean Distances}} &
        \textbf{HumanEval (\%)} \\
        \cmidrule(lr){2-4}
        & C--C & NC--NC & C--NC & \\
        \midrule
        \textbf{Llama2-7B}       & $1.4961 \pm 0.0060$ & $1.3741 \pm 0.0313$ & $1.4465 \pm 0.0148$ & 12.2 \\
        \textbf{CodeLlama-7B}    & $2.3417 \pm 0.0110$ & $2.3437 \pm 0.0475$ & $2.3623 \pm 0.0249$ & 33.5 \\
        \midrule
        \textbf{Llama2-13B}      & $2.1274 \pm 0.0098$ & $1.9002 \pm 0.0524$ & $2.0324 \pm 0.0247$ & 20.1 \\
        \textbf{CodeLlama-13B}   & $2.5499 \pm 0.0134$ & $2.5305 \pm 0.0539$ & $2.5597 \pm 0.0275$ & 36.0 \\
        \bottomrule
    \end{tabular}
    }
\end{table}

\begin{table}[h!]
    \centering
    \footnotesize
    \caption{\textbf{Embedding Dispersion vs.\ HumanEval Performance (Llama2 vs.\ CodeLlama, Cosine).}
    We report the mean $\pm$ standard error of \textbf{Cosine} distances among code tokens (C--C), among non-code tokens (NC--NC), and between code and non-code tokens (C--NC). We also list each model’s HumanEval pass@1 (\%).}
    \label{tab:llama-code-cosine-dispersion}
    \resizebox{0.88\linewidth}{!}{
    \begin{tabular}{lcccc}
        \toprule
        \textbf{Model} &
        \multicolumn{3}{c}{\textbf{Cosine Distances}} &
        \textbf{HumanEval (\%)} \\
        \cmidrule(lr){2-4}
        & C--C & NC--NC & C--NC & \\
        \midrule
        \textbf{Llama2-7B}       & $0.9603 \pm 0.0021$ & $0.9402 \pm 0.0075$ & $0.9652 \pm 0.0028$ & 12.2 \\
        \textbf{CodeLlama-7B}    & $0.9451 \pm 0.0027$ & $0.9486 \pm 0.0074$ & $0.9636 \pm 0.0037$ & 33.5 \\
        \midrule
        \textbf{Llama2-13B}      & $0.9340 \pm 0.0025$ & $0.9283 \pm 0.0101$ & $0.9460 \pm 0.0061$ & 20.1 \\
        \textbf{CodeLlama-13B}   & $0.9433 \pm 0.0027$ & $0.9435 \pm 0.0078$ & $0.9589 \pm 0.0037$ & 36.0 \\
        \bottomrule
    \end{tabular}
    }
\end{table}

\clearpage

\subsubsection{Robustness of Dispersion Correlation across Training Trajectories}
\label{app:trajectory-analysis}

In \S\ref{sec:embedding-dispersion-modelsel}, we demonstrated a strong correlation between the Dispersion Gap and model performance across different model families. To verify that this relationship is not an artifact of small sample sizes or specific model architectures, we conducted a high-resolution analysis of a single model's training trajectory.

\paragraph{Setup.}
We utilized the training checkpoints of the Olmo-7B model, sampling every 20,000 steps from step 120,000 to step 700,000. We exclude the initial 120,000 steps to bypass the 'burn-in' phase and focus on the stable training regime where representational geometry has consolidated. This yielded a set of $N=30$ distinct checkpoints. For each checkpoint, we calculated the Dispersion Gap ($\mathcal{G}$) using our standard protocol and compared it against the model's zero-shot performance on \textsc{GSM8K} (Math) and \textsc{HumanEval} (Code).

\begin{figure}[h]
    \centering
    \includegraphics[width=\textwidth]{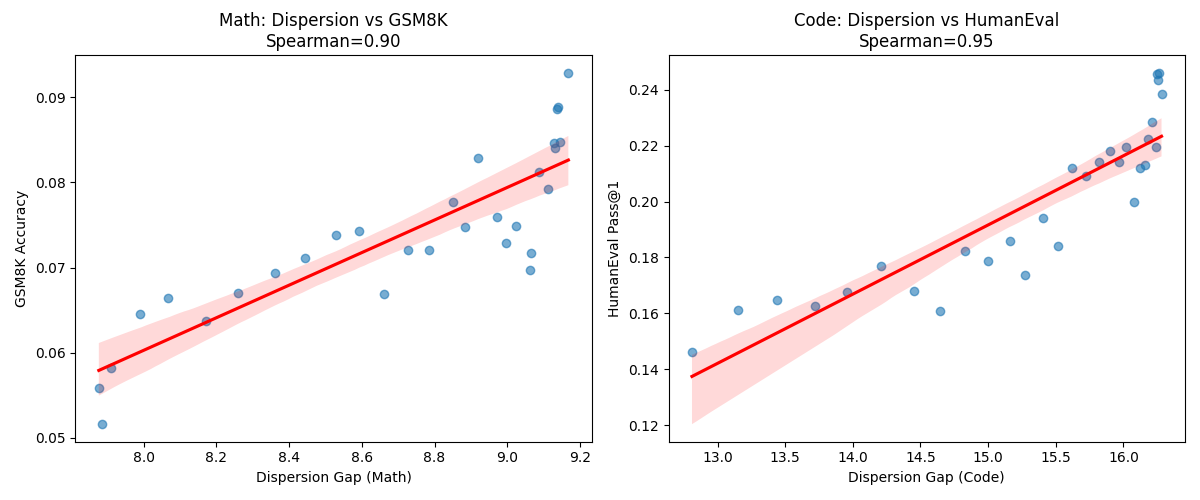} 
    \caption{\textbf{Dispersion Gap tracks performance across the Olmo-7B training trajectory ($N=30$).} 
    We observe extremely high Spearman rank correlations ($\rho \approx 0.90$ for Math and $\rho \approx 0.95$ for Code) with high statistical significance ($p < 10^{-11}$). Shaded regions indicate the 95\% confidence interval.}
    \label{fig:spearman_trajectory}
\end{figure}

\paragraph{Results.}
As shown in Figure~\ref{fig:spearman_trajectory}, we observe a remarkably strong monotonic relationship between dispersion and downstream accuracy throughout the training process. 
\begin{itemize}
    \item For Mathematical Reasoning, the correlation between the Dispersion Gap and \textsc{GSM8K} accuracy is $\rho=0.90$ ($p < 10^{-11}$).
    \item For Code Generation, the correlation between the Dispersion Gap and \textsc{HumanEval} pass@1 is $\rho=0.95$ ($p < 10^{-15}$).
\end{itemize}

These results reinforce our claim that representation dispersion serves as a robust, label-free proxy for model capability. Notably, the dispersion metric tracks the underlying improvement of the model with high fidelity, maintaining a strong correlation even amidst the natural fluctuations inherent in generation-based evaluation benchmarks.

\color{black}

\clearpage

\subsection{Details Regarding Layer Selection for kNN-LM}
\label{app:knn-layer-selection}

\subsubsection{Additional Results}

\begin{figure}[ht]
  \centering
  \includegraphics[width=0.9\linewidth]{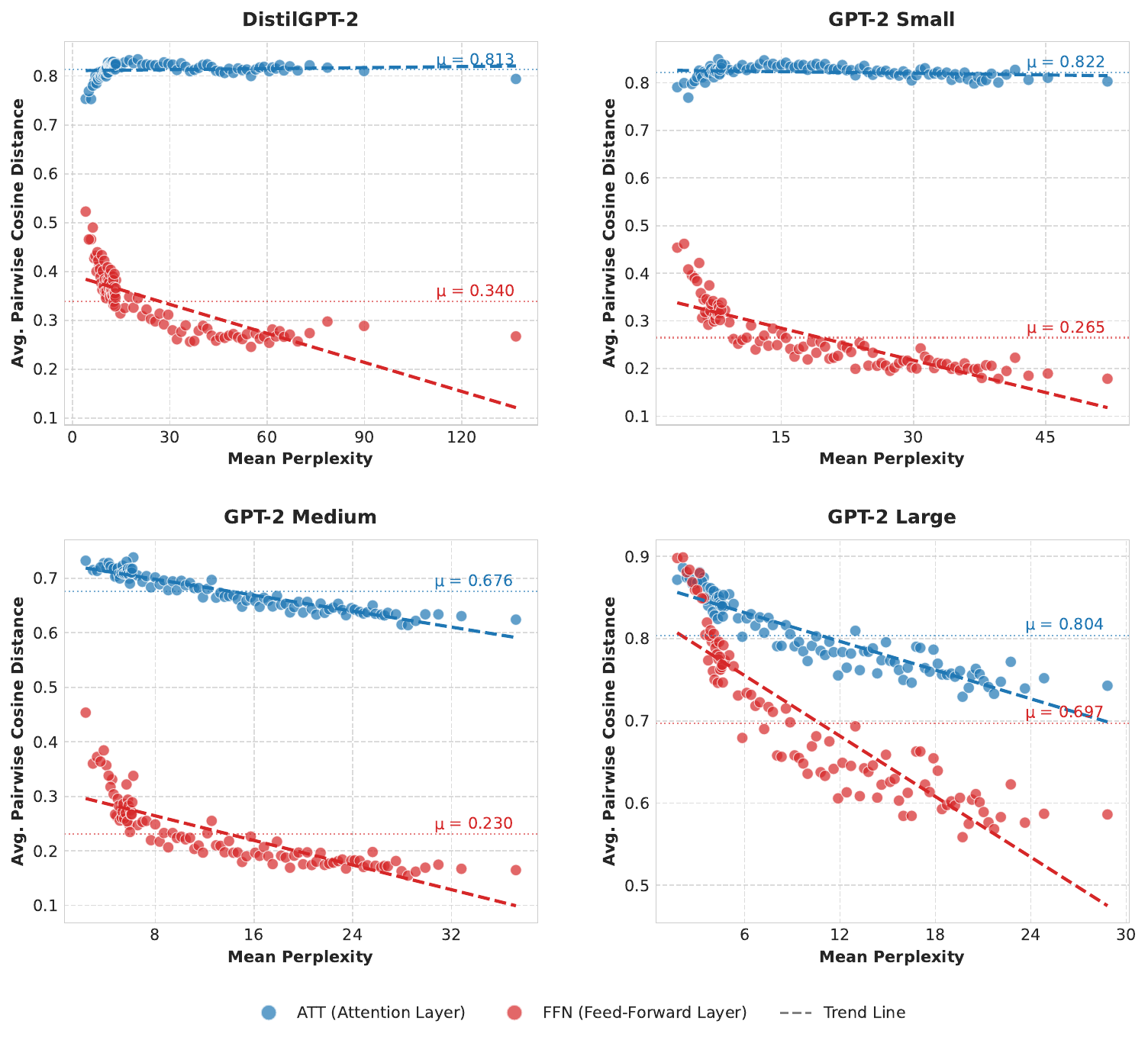}
  \caption{Mean perplexity vs.\ sub-layer average pairwise cosine distance for four 
  GPT-2 variants (\texttt{distilgpt2}, \texttt{gpt2}, \texttt{gpt2-medium}, 
  \texttt{gpt2-large}). Each point is a 512-token chunk of text.}
  \label{fig:knn_sent_ppl_full}
\end{figure}

We present extended findings on sub-layer selection for \(k\)NN-LM. 
Figure~\ref{fig:knn_sent_ppl_full} displays results for \emph{four} GPT-2 variants 
(\texttt{distilgpt2}, \texttt{gpt2}, \texttt{gpt2-medium}, \texttt{gpt2-large}). 
As in the main text, each point represents a 512-token chunk of text, with its 
mean perplexity plotted against the sub-layer’s average pairwise cosine distance 
(blue for the attention output, red for the feed-forward output). Interestingly, the negative correlation is weaker for the attention output than for the feed-forward output.

\clearpage

\subsection{Details Regarding Incorporating Representation Dispersion}
\label{app:training-details}

\subsubsection{Training Details}

All experiments in \S\ref{sec:rd-regularization} were conducted on NVIDIA A100 80GB GPUs.

\paragraph{Single-Domain Setting.} 
We train \textsc{gpt2-small} from scratch on WikiText, using a batch size of 64 and a block size (sequence length) of 512. The auxiliary loss weight $\lambda$ is tuned over the set:
\[
    \{0.5,\, 0.2,\, 0.1,\, 0.07,\, 0.05,\, 0.02,\, 0.01,\, 0.007,\, 0.005,\, 0.002,\, 0.001\}
\]
We experiment with learning rates $\{1 \times 10^{-3},\, 7 \times 10^{-4},\, 5 \times 10^{-4}\}$.

\paragraph{Cross-Domain Setting.} 
For joint WikiText + Python code training, we similarly use \textsc{gpt2-small} (from scratch), a batch size of 128, and a block size of 256. The auxiliary loss weight $\lambda$ and learning rates are swept over the same sets as above:
\begin{itemize}
    \item $\lambda$ values: $\{0.5,\, 0.2,\, 0.1,\, 0.07,\, 0.05,\, 0.02,\, 0.01,\, 0.007,\, 0.005,\, 0.002,\, 0.001\}$
    \item Learning rates: $\{1 \times 10^{-3},\, 7 \times 10^{-4},\, 5 \times 10^{-4}\}$
\end{itemize}

For both settings, we select $\lambda$ by validation for each learning rate. All experiments use standard AdamW optimizer settings unless otherwise specified.

\clearpage

\subsubsection{Impact of Enforcing Clusterization (``Squeeze'' Ablation)}
\label{app:squeeze-ablation}

To further investigate the causal relationship between representation dispersion and model performance, we conducted an ablation study where we explicitly discouraged dispersion. In contrast to the ``push-away'' objective described in \S\ref{sec:rd-regularization}, here we added an auxiliary loss that minimizes the average pairwise cosine distance, effectively squeezing representations into tighter clusters.

We trained \textsc{gpt2-small} on \textsc{Wikitext-103} using the same hyperparameter sweep as our baseline experiments. Table~\ref{tab:squeeze_results} compares the test set perplexity of the Baseline model against the ``Squeeze'' model (where the auxiliary loss weight $\lambda=0.1$).

\begin{table}[h!]
\centering
\small
\begin{tabular}{c c c c c}
\toprule
\multirow{2}{*}{\textbf{Learning Rate}} & \multirow{2}{*}{\textbf{Step}} & \multicolumn{2}{c}{\textbf{Perplexity (Lower is better)}} & \multirow{2}{*}{\textbf{$\Delta$ PPL}} \\
\cmidrule(lr){3-4}
 & & \textbf{Baseline} & \textbf{Squeeze (Cluster)} & \\
\midrule
\multirow{2}{*}{$1\times10^{-3}$}
 & 500 & 226.1 & 232.9 & +6.8 \\
 & 1000 & 111.3 & 137.0 & +25.7 \\
\midrule
\multirow{2}{*}{$7\times10^{-4}$}
 & 500 & 195.0 & 206.1 & +11.1 \\
 & 1000 & 96.7 & 124.2 & +27.5 \\
\midrule
\multirow{2}{*}{$5\times10^{-4}$}
 & 500 & 166.2 & 169.2 & +3.0 \\
 & 1000 & 83.0 & 101.7 & +18.7 \\
\bottomrule
\end{tabular}
\caption{\textbf{Effect of artificially reducing dispersion (``Squeeze'' ablation).}
Comparing the Baseline model to a model trained with an auxiliary loss that forces embeddings to cluster. Across all learning rates, restricting the geometry leads to significantly higher (worse) perplexity, reinforcing the causal link that adequate embedding breadth is necessary for strong predictive performance.}
\label{tab:squeeze_results}
\end{table}

The results show a consistent and significant degradation in performance when dispersion is penalized. For the optimal learning rate ($5\times10^{-4}$), the perplexity worsens by over 18 points at step 1000. This substantial performance drop supports the hypothesis that representation dispersion is a functional driver of model performance; when the model is geometrically constrained from spreading its representations, its ability to minimize entropy and predict accurately is directly impaired.

\clearpage
\subsubsection{Relation to Contrastive and Repulsive Objectives}
\label{app:contrastive}

Our auxiliary ``spread-out'' loss from \S\ref{sec:rd-regularization} is conceptually related to prior contrastive and repulsive objectives that encourage more discriminative or isotropic representations, rather than being an entirely new idea. For example, \citet{gunel2021supervisedcontrastivelearningpretrained} introduce a supervised contrastive loss for fine-tuning language models, in which representations of same-labeled examples are pulled together while representations of different classes are pushed apart. Likewise, \citet{jain-etal-2023-contraclm} propose ContraCLM, a contrastive framework for causal language models that explicitly improves the discrimination of token and sequence representations. Traditional contrastive losses (e.g., InfoNCE-style objectives) follow the same spirit: they add a term during training that repels embeddings from one another (except for designated positive pairs), mitigating representation collapse and improving downstream generalization.

Our formulation can be viewed as a deliberately simplified variant of these contrastive objectives. In contrast to InfoNCE-based losses, we do not define positive pairs or sample a specific set of negatives for each anchor. Instead, our loss considers \emph{all} pairs of hidden states in the batch and directly maximizes their average cosine distance. This simplicity has two practical benefits: (i) it avoids sampling bias and the need for careful negative mining—every pair contributes equally to the repulsive force, so we are less sensitive to mini-batch composition or heuristic data augmentations; and (ii) it is easy to implement alongside the standard language-modeling loss, involving only a batch-wise cosine-distance computation without large softmax denominators or momentum-queue mechanisms. In this sense, our ``push-away'' loss can be viewed as a minimal, unsupervised contrastive regularizer that encourages a more uniform spread of representations on the unit sphere.

\color{black}

\clearpage

\section{Limitations}
\label{sec:limitations}
While our findings underscore a strong empirical link between representation dispersion and model performance, there are several limitations. First, our analyses focus on average pairwise cosine distances of final-layer representations, which may not capture all nuanced aspects of embedding geometry or model behavior. Second, although we observe consistent negative correlations between dispersion and perplexity across several model families and domains, causality cannot be definitively concluded; certain architectures or objectives may modulate this relationship in unforeseen ways. Third, our experiments center primarily on English text from standard benchmarks and a limited set of specialized domains (e.g.\ code, scientific abstracts). It remains unclear how well our observations extend to other languages, modalities, or highly domain-specific corpora. Further research is needed to fully understand these trade-offs and develop robust methods for controlling embedding geometry.

\end{document}